\ifdefined\pdfoutput
  \pdfoutput=1
\fi
\documentclass{article}

\usepackage{times,natbib}
\setcitestyle{authoryear,round,citesep={;},aysep={,},yysep={;}}
\let\htmlEmailSize\footnotesize
\makeatletter
\renewcommand{\normalsize}{\@setfontsize\normalsize{10pt}{11pt}}
\renewcommand{\small}{\@setfontsize\small{9pt}{10pt}}
\renewcommand{\footnotesize}{\@setfontsize\footnotesize{9pt}{10pt}}
\renewcommand{\scriptsize}{\@setfontsize\scriptsize{7pt}{8pt}}
\renewcommand{\large}{\@setfontsize\large{12pt}{14pt}}
\renewcommand{\LARGE}{\@setfontsize\LARGE{17pt}{20pt}}
\normalsize
\renewcommand{\section}{\@startsection{section}{1}{\z@}{-2ex plus -0.5ex minus -0.2ex}{1.5ex plus 0.3ex minus 0.2ex}{\large\scshape\raggedright}}
\renewcommand{\subsection}{\@startsection{subsection}{2}{\z@}{-1.8ex plus -0.5ex minus -0.2ex}{0.8ex plus 0.2ex}{\normalsize\scshape\raggedright}}
\renewcommand{\subsubsection}{\@startsection{subsubsection}{3}{\z@}{-1.5ex plus -0.5ex minus -0.2ex}{0.5ex plus 0.2ex}{\normalsize\scshape\raggedright}}
\renewcommand{\maketitle}{\par\begingroup\centering{\LARGE\scshape\@title\par}\ifx\@author\@empty\else\vspace{12pt}{\normalsize\pdfauthorblock\par}\fi\endgroup\vspace{12pt}\setcounter{footnote}{0}}
\renewenvironment{abstract}{\vspace{0.075in}\centerline{\large\scshape Abstract}\vspace{0.5ex}\begin{quote}}{\par\end{quote}\vspace{1ex}}
\makeatother
\ifdefined\pdfoutput
  \usepackage[utf8]{inputenc}
\fi
\usepackage[T1]{fontenc}
\usepackage{hyperref,booktabs,array,multirow,adjustbox}
\newcommand{\tablestyle}[1][5pt]{\small\setlength{\tabcolsep}{#1}\renewcommand{\arraystretch}{1.12}}
\newcommand{\tableranking}{\textbf{Bold}: best; \underline{underlined}: second best by unrounded values in each column.}
\newcommand{\mainranking}{\textbf{Bold}: best; \underline{underlined}: second best by unrounded values within each context group and column.}
\usepackage{graphicx,subcaption,wrapfig,placeins,float}
\usepackage{xspace}
\usepackage[table]{xcolor}
\usepackage{microtype,tabularx,tikz}

\hypersetup{colorlinks=true,allcolors=black}

\newlength{\mazeRowGap}
\definecolor{ablationRecon}{HTML}{DFF9ED}
\definecolor{ablationMultiFreq}{HTML}{E3EBFF}
\definecolor{ablationLatent}{HTML}{FFE3E7}
\definecolor{analysisWindow}{HTML}{B6E6CF}   \definecolor{analysisHistory}{HTML}{BFCFF2}  \definecolor{analysisMovement}{HTML}{F2C9AD} \definecolor{analysisIndividuality}{HTML}{EDBFD0} \definecolor{analysisBehavior}{HTML}{D3BCE5}
\DeclareRobustCommand{\concepttag}[2]{\leavevmode
  \tikz[baseline=(concepttag.base)]{\node[fill=#1,rounded corners=1.6pt,inner xsep=2.3pt,inner ysep=1.2pt,text depth=.15ex,text height=1.65ex] (concepttag) {\strut #2};}}

\usepackage{amsmath,amsfonts}

\usepackage[nameinlink,noabbrev]{cleveref}

\title{Representing Short-Term Movement to Reveal Animal Identity and Learning Progress}
\author{
\begin{tabular}{ccc}
Ping~Zhang\textsuperscript{1,*} & Quang-Huy~Nguyen\textsuperscript{1,*} & Yingtie~Lei\textsuperscript{2,*}\\
\mbox{\htmlEmailSize\href{mailto:pingzh@bu.edu}{pingzh@bu.edu}} & \mbox{\htmlEmailSize\href{mailto:quanghuy@bu.edu}{quanghuy@bu.edu}} & \mbox{\htmlEmailSize\href{mailto:lei.441@osu.edu}{lei.441@osu.edu}}\\
\rule{0pt}{10pt} & & \\
Sameera~Shridhar\textsuperscript{3} & Zachary~Leveroni\textsuperscript{3} & Ellie~Quattrocchi\textsuperscript{3}\\
\mbox{\htmlEmailSize\href{mailto:sameerashridhar@uchicago.edu}{sameerashridhar@uchicago.edu}} & \mbox{\htmlEmailSize\href{mailto:leveroniz@uchicago.edu}{leveroniz@uchicago.edu}} & \mbox{\htmlEmailSize\href{mailto:equattrocchi@uchicago.edu}{equattrocchi@uchicago.edu}}\\
\end{tabular}
\\[8pt]
\begin{tabular}{ccc}
Jai~Y~Yu\textsuperscript{3} & \rule{50pt}{0pt} & Wei-Lun~Chao\textsuperscript{1,2}\\
\mbox{\htmlEmailSize\href{mailto:jaiyu@uchicago.edu}{jaiyu@uchicago.edu}} & & \mbox{\htmlEmailSize\href{mailto:chao209@bu.edu}{chao209@bu.edu}}\\
\end{tabular}
\\[11pt]
\textsuperscript{1}Boston University\quad\textsuperscript{2}The Ohio State University\quad\textsuperscript{3}University of Chicago\\[3pt]
\mbox{\textsuperscript{*}Equal contribution.}
}
\newcommand{\pdfauthorblock}{
  \large
  \makebox[.32\textwidth]{Ping~Zhang\textsuperscript{1,*}}\makebox[.32\textwidth]{Quang-Huy~Nguyen\textsuperscript{1,*}}\makebox[.32\textwidth]{Yingtie~Lei\textsuperscript{2,*}}\\[0pt]
  {\small\rmfamily\color{black!72}\makebox[.32\textwidth]{pingzh@bu.edu}\makebox[.32\textwidth]{quanghuy@bu.edu}\makebox[.32\textwidth]{lei.441@osu.edu}}\\[8pt]
  \makebox[.32\textwidth]{Sameera~Shridhar\textsuperscript{3}}\makebox[.32\textwidth]{Zachary~Leveroni\textsuperscript{3}}\makebox[.32\textwidth]{Ellie~Quattrocchi\textsuperscript{3}}\\[0pt]
  {\small\rmfamily\color{black!72}\makebox[.32\textwidth]{sameerashridhar@uchicago.edu}\makebox[.32\textwidth]{leveroniz@uchicago.edu}\makebox[.32\textwidth]{equattrocchi@uchicago.edu}}\\[8pt]
  \makebox[.32\textwidth]{Jai~Y~Yu\textsuperscript{3}}\makebox[.32\textwidth]{Wei-Lun~Chao\textsuperscript{1,2}}\\[0pt]
  {\small\rmfamily\color{black!72}\makebox[.32\textwidth]{jaiyu@uchicago.edu}\makebox[.32\textwidth]{chao209@bu.edu}}\\[11pt]
  {\normalsize\color{black!85}\mbox{\textsuperscript{1}Boston University\enspace\textsuperscript{2}The Ohio State University\enspace\textsuperscript{3}University of Chicago}}\\[3pt]
  {\normalsize\color{black!72}\textsuperscript{*}Equal contribution.}
}
\date{}

\newcommand{\ours}{\textsc{TRAIL}\xspace}
\newcommand{\ourslong}{TRAjectory-Identity Latent\xspace}

\newcommand{\resultfigure}[4][!htbp]{
  \begin{figure}[#1]
    \centering
    #2
    \caption{#3}
    \label{#4}
  \end{figure}
}

\newcommand\mypara[1]{\par\addvspace{0.6mm}\noindent\strut\textbf{#1}\enspace\ignorespaces}
\newcommand{\needroom}[1]{\par\begingroup\ifdim\dimexpr\pagegoal-\pagetotal\relax<#1\relax\newpage\fi\endgroup}

\begin{document}
  \maketitle
  \addtocontents{toc}{\protect\setcounter{tocdepth}{-1}}

  \begin{abstract}
    How learning progresses over time and varies across individuals is a longstanding question in cognitive and behavioral science. Behavioral changes are observable outcomes of learning and have been a long-established bedrock for understanding learning progress. However, a major challenge is identifying observable behavioral measures that faithfully capture learning progression. Existing approaches commonly rely on predefined behavioral features or task-level performance, while fine-scale posture and movement in freely behaving animals remain less explored. In this paper, we ask: \emph{can short-term movement trajectories reveal learning progress and individual identity?} We introduce a self-supervised representation learning framework built on a simple hypothesis: behavioral characteristics associated with identity and learning remain relatively stable over nearby time intervals. Using temporal proximity as supervision, our framework learns compact representations directly from movement trajectories without manually defining behavioral states or kinematic features. Using behavior data from freely behaving rats, we show that the learned representations not only improve prediction of identity and learning-related readouts over existing methods, but also provide a tool for investigating broader scientific questions about when and how these signals appear in behavior. Specifically, trajectories as short as one second contain informative learning and identity signals. Egocentric pose dynamics are more predictive of identity, whereas allocentric motion is more predictive of learning-related variation, and both signals are strongest at decision junctions in the alternation cohort. We further use a representation-conditioned diffusion model to inspect movement patterns associated with identity and learning progress. Our findings establish short-term movement as an informative behavioral ``fingerprint'' for studying learning and individual differences.
    Our code is available at \href{https://github.com/pizhn/TRAIL}{github.com/pizhn/TRAIL}.
  \end{abstract}

  \section{Introduction}

  A longstanding question in cognitive science is how learning progresses over time. We can study this progression by characterizing changes in brain networks and behavior. While characterizing neural changes requires recording brain activity, behavioral changes provide a directly observable window into the learning process. A major challenge, however, is identifying which aspects of behavior show learning-related changes and how they evolve. Standard approaches rely on predefined behavioral features, such as speed, latency, or the frequency of a target behavior, as well as task-level performance measures, such as success rate. These measures are interpretable and readily quantifiable, but often summarize behavior as coarse scalar changes over extended periods~\citep{smith2004dynamic}. In contrast, fine-scale changes in limb posture and movement can provide a wealth of information about learning progression, as demonstrated in studies of motor learning~\citep{kawai2015motor,bova2021reaching,dhawale2019adaptive,nguyen2021kinematic}. Yet, in unconstrained settings, where animals learn to navigate freely in space~\citep{rosenberg2021labyrinth}, the complexity of movement patterns makes it difficult to determine which behavioral features to measure and how they can be used to understand changes in the underlying cognitive processes. Addressing this challenge could reveal the hidden structures and dynamics of learning.

  In addition to behavioral changes over time, inter-individual differences in movement have been reported across species~\citep{godesberg2024walking,chao2019gaitset}. Such movement ``fingerprints'' are increasingly being linked to biologically important variation in genetic and cognitive traits~\citep{levy2023individual,forkosh2019identity,freund2013individuality,vonziegler2024flow,sheppard2022stride,ding2026geese}. If successfully applied to learning-related movement analysis, these fingerprints could add a new dimension to understanding individual learning dynamics. However, current methods vary widely in their performance and the observation periods required to identify individuals, ranging from tens of seconds to minutes~\citep{levy2023individual,forkosh2019identity,vonziegler2024flow}.

  These limitations raise a fundamental question: \emph{can short-term movement trajectories reveal learning progress and individual identity?} We focus on body posture and movement patterns unfolding over only a few seconds---a timescale that is often insufficient to assess task-level performance. Nevertheless, how animals move, coordinate their body parts, or respond to their surroundings within these short intervals may already provide informative signals~\citep{weinreb2026tasks}. We investigate whether a representation learned directly from movement trajectories can capture these signals without manually defining behavioral states or kinematic features.

  \begin{figure}[!t]
    \centering
    \includegraphics{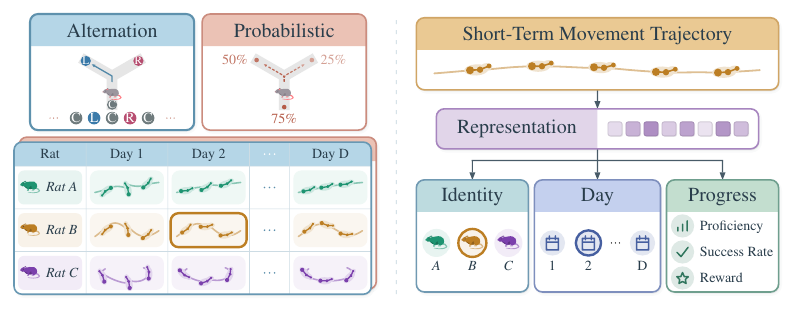}
    \vspace{-5pt}
    \caption{Dataset and problem setting. \textit{Left:} Repeated pose recordings span animals and days under two reward rules over the Y-shape spatial learning dataset: alternation and probabilistic rewards. L, C, and R denote the left, center, and right arms, respectively. \textit{Right:} A compact representation of short-term movement trajectories is learned and evaluated for inferring identity, recording day, and performance (e.g., proficiency, success rate, and reward).}
    \label{fig:teaser}
    \vspace{-15pt}
  \end{figure}

  To this end, we study movement data from rat spatial learning experiments described in~\citep{ShridharENEURO.0048-26.2026}. We first evaluate existing approaches for representing movement trajectories, spanning traditional state-based methods~\citep{Durbin_2012,Wiltschko_2015,NIPS2013_e53a0a29,Weinreb2024}, recent machine learning methods~\citep{Hinton_2006,Luxem2022,NEURIPS2020_f3ada80d,BaiTCN2018,NEURIPS2023_5aad86aa,Stoffl2025hBehaveMAE}, and time-series foundation models~\citep{google2026timesfm3,ansari2025chronos2}. Across identity- and learning-related evaluations such as task performance, these approaches have limited predictive performance.

  We therefore develop a representation learning framework, \ours (\ourslong), tailored to capturing information about individual differences and learning progress from short segments of movement data, as illustrated in ~\cref{fig:teaser}. Our framework combines a trajectory encoder with a self-supervised objective. Inspired by Wayformer~\citep{nayakanti2022wayformer}, an architecture developed for short-term motion forecasting in autonomous driving, the encoder maps short movement trajectories into compact latent representations. Our training objective builds on a simple hypothesis: although an animal's movements may vary rapidly, the underlying behavioral characteristics associated with identity and learning are more stable over nearby time intervals. We therefore use temporal proximity as a self-supervisory signal, encouraging representations of nearby trajectory segments to be similar while contrasting them with segments from other recordings or sufficiently distant times within the same recording. This objective enables representation learning without identity or learning-progress labels during training.

  We evaluate our framework on the spatial learning data~\citep{ShridharENEURO.0048-26.2026} and find clear improvements over existing methods. Linear probes trained on the learned representations predict learning-related readouts more accurately, demonstrating that such information is accessible from short movement trajectories. Interestingly, the representations also support accurate identity prediction, even though segments from different recordings of the same individual can be contrasted during training. Beyond these learning-focused datasets, our method remains competitive on the established behavioral benchmark MABe22~\citep{pmlr-v202-sun23g}, suggesting broader applicability to representation learning from short movement sequences.

  The superior performance of our framework not only provides an affirmative answer to our fundamental question, but also offers a tool for investigating broader scientific questions: \emph{At what timescale does behavior vary in relation to learning? How long must a trajectory be to capture identity- and learning-related signals, and under what conditions are these signals strongest? Which movement patterns reveal individual identity and learning progress?}

  First, our temporal-window analysis favors local temporal neighborhoods: expanding the positive-pair window beyond the default 10\% of a recording partition reduces prediction of learning-related readouts. Second, trajectories as short as one second already contain information predictive of both identity and learning-related readouts. In the alternation cohort, both signals are stronger at locations where animals must choose a direction than along straight paths. Third, local pose dynamics are more predictive of identity, whereas global motion is more predictive of learning-related readouts. Together, these findings identify temporal, spatial, and postural factors that can guide more targeted hypotheses about behavior and learning. Finally, we use a representation-conditioned diffusion model~\citep{peebles2023scalable} to explore the movement patterns associated with individual identity and learning progress. By varying the conditioning representation, we generate trajectories that preserve broad motion structure while exhibiting subtle changes in stopping, accelerating, and stretching. These qualitative results make behavioral differences associated with the learned representations directly inspectable.

  Taken together, our study shows that short movement trajectories contain observable signatures of both individual identity and learning progress. By combining self-supervised representation learning with controlled analyses and conditional generation, we provide a framework for predicting these properties and examining the movement patterns associated with them.

  \section{Related Work and Background}

  \mypara{Behavioral tasks \& datasets.}
  CalMS21~\citep{calms21} and MABe22~\citep{pmlr-v202-sun23g} support behavioral recognition and representation learning across diverse settings, while longitudinal resources such as CRCNS hc-28~\citep{crcns_hc28} track rats learning rewarded alternation across days. In this paper, we use data from two spatial navigation tasks for rats~\citep{ShridharENEURO.0048-26.2026}. The alternation and probability reward rules define learning objectives with quantifiable performance metrics over days. Unlike spontaneous behavior, this design allows us to relate locomotion and posture features to learning progress and experience in the environment. Longitudinal recordings of the same animals over consecutive days enable analysis of individual learning trajectories, with body keypoint coordinates available for both conditions. The two reward rules provide complementary evaluation settings, including behavior under reward uncertainty.

  \mypara{Algorithms.} Reconstruction, prediction, and representation alignment provide approaches to learning behavioral structure without labels. VAME~\citep{Luxem2022} combines reconstruction with future prediction, while BAMS~\citep{NEURIPS2023_5aad86aa} and hBehaveMAE~\citep{Stoffl2025hBehaveMAE} capture behavioral structure at multiple temporal scales. Temporal Neighborhood Coding uses local stationarity to define temporal neighborhoods and learns to distinguish neighboring from non-neighboring segments with a debiased contrastive objective~\citep{tonekaboni2021tnc}. CEBRA supports both label-free temporal sampling and sampling guided by auxiliary variables~\citep{schneider2023cebra}. Building on temporal supervision, our approach combines contrastive learning with reconstruction to learn a compact representation from short trajectory windows. This representation supports analysis of movement changes during learning and differences across individuals.

  \section{Proposed Representation Learning Framework: \ours}
  We learn compact representations of short movements without predefined behavioral categories or hand-crafted features, and evaluate their ability to capture identity and learning-related variation.

  \subsection{Overview}

  \mypara{Data format.} We use data recorded from each animal over multiple days, with multiple sessions per day~\citep{ShridharENEURO.0048-26.2026}. Each trajectory segment $x_i\in\mathbb{R}^{T\times K\times2}$ contains $T$ frames of $K$ keypoints and is associated with animal $r_i$, session $s_i$, start time $t_i$, and within-animal recording day $d_i$.
  We consider three views of movement: local pose, global trajectory, and full movement (\cref{app:traject_view}).
  We denote the selected view by $v_i=V(x_i)$ and optional environmental geometry by $c_i$.

  \begin{figure}[t]
    \centering
    \includegraphics{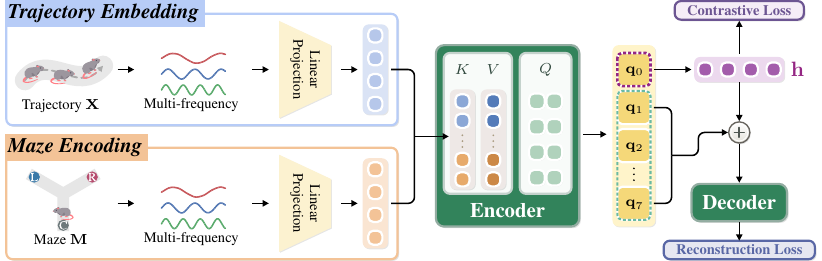}
    \caption{\ours high-level architecture.}
    \label{fig:main_method}
  \end{figure}

  \mypara{Design principles.} To capture identity-specific patterns and learning-related changes from short-term movements, we consider three design choices: (1) multi-frequency encoding to preserve fine-grained movement details~\citep{mildenhall2022nerf}, (2) a compact latent space learned through contrastive learning, based on the hypothesis that nearby trajectory segments share similar underlying behavioral states, and (3) an auxiliary reconstruction objective. Together, these choices encourage \ours to retain informative movement details and learn behavioral relationships without predefined categories or explicit labels.

  \mypara{High-level architecture.} \Cref{fig:main_method} shows the encoder-decoder architecture of \ours. Inspired by Wayformer~\citep{nayakanti2022wayformer}, the encoder uses learnable queries to summarize multi-frequency embeddings of movement trajectories and environmental context (e.g., maze layout). The first output token is compressed into a behavioral representation trained through contrastive learning, and then projected back to the original feature dimension and combined with the remaining tokens for trajectory reconstruction. Reconstruction encourages the model to retain fine-grained movement details. We detail each component below.

  \subsection{Model Architecture and Learning Objectives}
  \label{app:architecture_detail}
  \mypara{Multi-frequency trajectory embedding.} We take keypoint trajectories $\mathbf{X}\in\mathbb{R}^{T\times K\times2}$ and maze coordinates $\mathbf{M}\in\mathbb{R}^{N\times2}$ as input; the maze provides global context for the trajectory. Inspired by Wayformer~\citep{nayakanti2022wayformer}, we anchor each trajectory to its final frame by centering the body point and vertically aligning the body-neck vector in that frame. We then apply the same transformation to all trajectory frames and maze coordinates. Each two-dimensional coordinate receives NeRF-style multi-frequency encoding with $L=10$ frequencies~\citep{mildenhall2022nerf} and a linear projection to 64 dimensions.
  To identify each token's corresponding keypoint and time frame, we introduce two learnable positional embeddings: a keypoint embedding shared across frames and a temporal embedding shared across keypoints within the same frame.
  Maze coordinates undergo the same encoding procedure, with a learnable additive embedding distinguishing them from trajectory tokens.
  Together, the $K\times T$ trajectory tokens and $N$ maze tokens form $\mathbf{Z}\in\mathbb{R}^{(KT+N)\times64}$.

  \mypara{Compact behavioral representation.}
  We adopt a four-layer Q-Former~\citep{li2023blip2} as the encoder, which takes eight learned queries to attend to $\mathbf{Z}$, yielding $\mathbf{Q}=[\mathbf{q}_0,\ldots,\mathbf{q}_7]\in\mathbb{R}^{8\times64}$. We then linearly project $\mathbf{q}_0$ to produce the 16-dimensional latent behavior representation $\mathbf{h}$.
  To learn the representations, we hypothesize that nearby trajectory segments share similar underlying behavioral states.
  We therefore use a balanced InfoNCE loss~\citep{oord2018representation}. Positive pairs consist of segments from the same session within a certain time interval; negatives are drawn from temporally distant segments or different sessions:
  \begin{equation}
    \label{eq:contrastive}
    \mathcal{L}_{\mathrm{contrast}}=-\frac{1}{|\mathcal{P}|}\sum_{\mathbf{h}^{+}\in\mathcal{P}}\log\frac{|\mathcal{P}|^{-1}\exp\!\big(\ell(\mathbf{h},\mathbf{h}^{+})\big)}{|\mathcal{P}|^{-1}\sum_{\mathbf{h}'\in\mathcal{P}}\exp\!\big(\ell(\mathbf{h},\mathbf{h}')\big)+|\mathcal{N}|^{-1}\sum_{\mathbf{h}^{-}\in\mathcal{N}}\exp\!\big(\ell(\mathbf{h},\mathbf{h}^{-})\big)}.
  \end{equation}
  where $\mathcal{P}$ and $\mathcal{N}$ denote the positive and negative sets for an anchor representation $\mathbf{h}$, and $\ell(\mathbf{h},\mathbf{h}')=s(\mathbf{h},\mathbf{h}')/\tau$ is cosine similarity scaled by temperature $\tau=0.1$. The compressed representation $\mathbf{h}$ is then projected back to 64 dimensions and concatenated with the remaining tokens $q_{1:7}$ to form the decoder input for trajectory reconstruction.

  \mypara{Decoder for reconstruction.} The decoder follows the same Q-Former architecture as the encoder, with separate parameters and learnable queries that attend to the encoder output. An MLP head maps the first query representation to the parameters of a Gaussian distribution over the trajectory, which we train using a negative log-likelihood reconstruction loss $\mathcal{L}_{\mathrm{recon}}$.

  \mypara{Overall objective.} The final objective weights contrastive learning and reconstruction equally:
  \begin{equation}
    \mathcal{L}_{\mathrm{total}} = \mathcal{L}_{\mathrm{contrast}} + \mathcal{L}_{\mathrm{recon}}.
  \end{equation}
  We evaluate each component through ablation studies in \cref{app:ablation}.

  \section{Experiments on Representation Learning}
  \label{sec:experiments}

  Our experiments proceed from prediction to scientific analysis and interpretation. We first compare frozen representations on identity and learning-related readouts in the two Y-maze Spatial learning cohorts, then evaluate their applicability on MABe22 Mouse Triplets. We next examine temporal settings, separate local pose from global movement, visualize representation structure, and compare signals across maze regions, recording days, and movement motifs. Finally, we use conditional generation to inspect movement changes associated with progression across days and variation between individuals. The appendix provides implementation details, ablations, robustness and generalization tests, and additional visualizations.

  \mypara{Dataset description.}
  We use data from two spatial-learning cohorts recorded in a Y-shaped maze \citep{ShridharENEURO.0048-26.2026}: six rats learning an \textit{alternation} rule that rewards goal visits in a prescribed sequence, and eight different rats learning a \textit{probabilistic} rule with goal-specific reward probabilities. The maze consists of three goal platforms connected to a central junction by arms. We represent its geometry using $25$ waypoints.
  Recordings span five consecutive days with 3--4 approximately ten-minute sessions per day at 11 Hz, totaling 275 sessions and 44.92 hours. We consider the nose, neck, and body keypoints, which are the most reliably tracked using DeepLabCut~\citep{mathis2018deeplabcut}.
  We split each session chronologically into 80\% training, 5\% validation, and 15\% testing, and then aggregate them across sessions to form the training, validation, and test sets (\cref{app:maze_dataset_statistics}).
  \mypara{Evaluation metrics.}
  We consider five evaluation targets: \textit{Identity}, \textit{Day}, trial-level \textit{Proficiency}, and \textit{Success Rate} and \textit{Reward} over one-minute intervals. Binary reward outcomes are recorded for each trial that spans departure from one goal platform to arrival at the next; we then obtain the ground-truth proficiency using the state-space model of \citet{smith2004dynamic}.

  Unless stated otherwise, we freeze the encoder and train a separate linear probe for each target. For the spatial-learning cohorts, we use one pooled logistic-regression probe for identity and a separate ridge-regression probe per animal for each continuous target. The MABe22 benchmark uses its own evaluation protocol (\cref{app:mabe_evaluation}). Probes are trained on the training trajectories and evaluated on the test trajectories. We report classification accuracy and mean squared error (MSE) for the regression tasks.

  \mypara{Baselines.}
  We compare three baseline groups. \textit{Classic} methods include linear probing on raw trajectories, Linear-Gaussian State-space Model~\citep{Durbin_2012}, Autoregressive Hidden Markov Model~\citep{Wiltschko_2015}, Probabilistic Movement Primitives~\citep{NIPS2013_e53a0a29}, and Keypoint-MoSeq~\citep{Weinreb2024}. \textit{Deep Learning} methods include AutoEncoder~\citep{Hinton_2006}, VAME~\citep{Luxem2022}, Temporal ConvNet~\citep{BaiTCN2018}, temporal BYOL~\citep{NEURIPS2020_f3ada80d}, BAMS~\citep{NEURIPS2023_5aad86aa}, and hBehaveMAE~\citep{Stoffl2025hBehaveMAE}. We also evaluate time-series \textit{Foundation Models}, Chronos-2~\citep{ansari2025chronos2} and TimesFM-3~\citep{pmlr-v235-das24c,google2026timesfm3}, frozen or with LoRA fine-tuning~\citep{hu2022lora}.

  We also consider a \textit{Supervised} baseline that uses the same trajectory embedding and encoder architecture as \ours but is trained from scratch jointly on all five targets (Identity, Day, Proficiency, Success Rate, and Reward), using cross-entropy for classification and MSE for regression.

  \mypara{Implementation details.}
  For the short-context Y-maze comparison, all methods use 3-second trajectories (i.e., $T = 3\times 11$), with BAMS restricted to its short-term branch. For each method, including \ours, we evaluate tens of hyperparameter configurations and select the best based on validation performance on Identity and Day in the alternation reward setting (\cref{app:implementation_detail}).

  \subsection{Results}

  \begin{table}[htbp]
    \centering
    \tablestyle
    \caption{Performance on the spatial-learning dataset (alternation rewards). \mainranking}
    \label{tab:yshape_alternation}
    \begin{adjustbox}{max width=\linewidth}
      \begin{tabular}{llccccc}
        \toprule
        & \multicolumn{1}{c}{\multirow{2}{*}{\textbf{Method}}} & \textbf{Identity} & \textbf{Day} & \textbf{Proficiency} & \textbf{Success Rate} & \textbf{Reward} \\
        & & (Acc $\uparrow$) & (MSE $\downarrow$) & (MSE $\downarrow$) & (MSE $\downarrow$) & (MSE $\downarrow$) \\
        \midrule
        \multirow{5}{*}{\textbf{Classic}} & Direct Trajectory & 27.16 & 1.96 & .021 & 0.14 & 0.76 \\
        & LG-SSM~\citep{Durbin_2012} & 26.05 & 2.02 & .021 & 0.14 & 0.77 \\
        & AR-HMM~\citep{Wiltschko_2015} & 44.09 & 3.69 & .023 & 0.21 & 0.69 \\
        & ProMPs~\citep{NIPS2013_e53a0a29} & 39.80 & 1.94 & .021 & 0.14 & 0.76 \\
        & Keypoint-MoSeq~\citep{Weinreb2024} & 27.43 & 1.98 & .021 & 0.14 & 0.77 \\
        \midrule
        \multirow{4}{*}{\textbf{Foundation Model}} & TimesFM-3 (Zero-shot)~\citep{google2026timesfm3} & 37.23 & 1.68 & .019 & 0.14 & 0.71 \\
        & TimesFM-3 (LoRA) & 44.41 & 1.64 & .019 & 0.14 & 0.68 \\
        & Chronos-2 (Zero-shot)~\citep{ansari2025chronos2} & 37.63 & 1.72 & .019 & 0.14 & 0.72 \\
        & Chronos-2 (LoRA) & 46.83 & 1.61 & .018 & 0.14 & 0.68 \\
        \midrule
        \multirow{8}{*}{\textbf{Deep Learning}} & AutoEncoder~\citep{Hinton_2006} & 59.45 & 1.61 & .018 & 0.13 & 0.68 \\
        & VAME~\citep{Luxem2022} & 62.51 & 1.61 & \underline{.017} & \underline{0.13} & 0.68 \\
        & T-BYOL~\citep{NEURIPS2020_f3ada80d} & 59.63 & 1.68 & .018 & 0.13 & 0.70 \\
        & TCN~\citep{BaiTCN2018} & 65.43 & \underline{1.51} & .018 & 0.14 & 0.70 \\
        & BAMS~\citep{NEURIPS2023_5aad86aa} & 63.74 & 1.59 & .018 & 0.13 & 0.67 \\
        & hBehaveMAE~\citep{Stoffl2025hBehaveMAE} & 47.81 & 1.64 & .018 & \textbf{0.13} & \underline{0.64} \\
        & Supervised & \underline{78.41} & 1.69 & .018 & 0.16 & 0.80 \\
        \rowcolor[HTML]{E3ECF5}       \cellcolor{white} & \ours & \textbf{84.00} & \textbf{1.26} & \textbf{.016} & 0.13 & \textbf{0.61} \\
        \bottomrule
      \end{tabular}
    \end{adjustbox}
  \end{table}

  \mypara{Spatial learning dataset.}
  \Cref{tab:yshape_alternation} shows that \ours outperforms all baselines on Identity and most learning-related tasks, except for Success Rate under the alternation reward rule. Its advantage over the Supervised baseline suggests that temporal contrastive learning and reconstruction encourage more effective representations than direct supervision alone. \ours also outperforms fine-tuned Chronos-2 and TimesFM-3, highlighting the value of representations tailored to movement trajectories. Similar trends hold under the probabilistic reward rule (\cref{app:alternation_results}).

  \mypara{Mouse Triplets dataset.}
  We also evaluate \ours on MABe22 Mouse Triplets~\citep{pmlr-v202-sun23g}, a behavioral dataset that does not focus on learning, comprising 5,336 one-minute clips and 13 tasks (4 sequence-level and 9 frame-level).

  \needroom{27\baselineskip}
  \begin{wraptable}{r}{0.6\textwidth}
    \centering
    \scriptsize
    \setlength{\tabcolsep}{2pt}
    \renewcommand{\arraystretch}{1.08}
    \caption{Performance on the MABe22 Mouse Triplets dataset with short- and long-term contexts. MSE values scaled by $100$.}
    \label{tab:mabe22}
    \begin{tabularx}{\linewidth}{ll*{5}{>{\centering\arraybackslash}X}}
      \toprule
      & \textbf{Method} & \textbf{Identity} & \textbf{Tasks} & \textbf{Tasks} & \textbf{Frame} & \textbf{Seq.} \\
      & & Acc$\uparrow$ & MSE$\downarrow$ & F1\%$\uparrow$ & F1\%$\uparrow$ & F1\%$\uparrow$ \\
      \midrule
      \multicolumn{7}{c}{\rule{0pt}{2.3ex}\textit{Short-term Context (2\,s)}} \\
      \midrule
      \multirow{4}{*}{\textbf{Classic}} & LG-SSM & 65.32 & 9.42 & 21.01 & 13.17 & 56.25 \\
      & AR-HMM & 45.62 & 9.36 & 22.50 & 14.70 & 57.61 \\
      & ProMPs & 67.65 & 9.43 & 22.59 & 14.73 & 57.96 \\
      & Keypoint-MoSeq & 63.50 & 9.39 & 22.42 & 13.67 & 61.79 \\
      \midrule
      \multirow{7}{*}{\textbf{Deep Learning}} & AutoEncoder & 67.53 & 9.39 & 25.16 & 17.06 & 61.62 \\
      & VAME & 68.24 & 9.36 & 25.93 & \underline{18.06} & 61.34 \\
      & T-BYOL & 69.30 & 9.43 & 25.72 & 17.13 & 64.36 \\
      & TCN & 67.26 & 9.45 & 24.72 & 16.00 & 63.97 \\
      & BAMS & 63.08 & 9.34 & 26.02 & 16.97 & 66.77 \\
      & hBehaveMAE & \underline{73.07} & \underline{9.31} & \textbf{29.88} & \textbf{20.78} & \underline{70.85} \\
      \rowcolor[HTML]{E3ECF5}       \cellcolor{white} & \ours & \textbf{73.65} & \textbf{9.26} & \underline{28.64} & 17.81 & \textbf{77.39} \\
      \midrule
      \multicolumn{7}{c}{\rule{0pt}{2.3ex}\textit{Long-term Context (45\,s)}} \\
      \midrule
      \multirow{3}{*}{\textbf{Deep Learning}} & BAMS & -- & \underline{9.24} & 27.74 & 17.36 & \underline{74.47} \\
      & hBehaveMAE & -- & 9.32 & \textbf{29.13} & \textbf{20.05} & 70.02 \\
      \rowcolor[HTML]{E3ECF5}       \cellcolor{white} & \ours & -- & \textbf{9.19} & \underline{29.02} & \underline{17.90} & \textbf{79.08} \\
      \bottomrule
    \end{tabularx}
  \end{wraptable}

  Following BAMS and hBehaveMAE, we use inductive evaluation with short (2\,s) and long (45\,s) contexts. We additionally evaluate within-clip \textit{Identity} by training a linear probe on the first 42 seconds and testing on the final 18 seconds. We omit this task in the long-context setting because the temporal splits are shorter than the context window. We use the reported configurations for BAMS and hBehaveMAE and those selected in our main experiments for the other baselines.
  \Cref{tab:mabe22} shows that \ours remains competitive at both context lengths and achieves the highest Identity accuracy. Its gains are stronger on sequence-level tasks than on frame-level tasks. These results demonstrate the applicability of \ours across experimental settings and timescales.
  \par\WFclear

  \section{Scientific Analysis and Insight}
  \label{sec:scientific-analysis}
  Having established strong identity and learning-related readouts across two datasets, we use \ours to investigate broader scientific questions: \emph{At what timescale does behavior vary in relation to learning? How long must a trajectory be to capture identity- and learning-related signals, and under what conditions are these signals strongest? Which movement patterns reveal individual identity and learning progress?} We conduct systematic, controlled analyses using the six rats performing a Y-maze alternation task. These analyses move beyond predictive performance to explore how brief movements reveal individuality and learning-related changes.

  \vspace{0.5em}

  \begin{figure}[!htbp]
    \centering
    \vskip-10pt
    \makebox[\linewidth][c]{
      \includegraphics{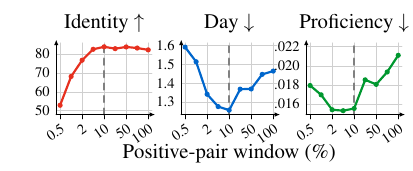}    \hfil
      \includegraphics{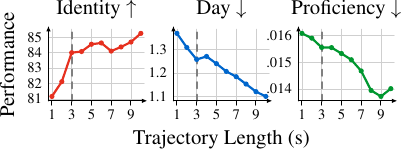}
    }
    \caption{Effects of temporal settings. \textit{Left:} Positive-pair window width, expressed as a percentage of the recording partition. \textit{Right:} Trajectory length. Dashed lines mark the default settings.}
    \label{fig:hist_len}
    \label{fig:positive_range}
  \end{figure}

  \mypara{\concepttag{analysisWindow}{At what timescale does behavior vary in relation to learning?}}
  Our contrastive learning objective assumes that learning unfolds gradually over hours to days~\citep{ShridharENEURO.0048-26.2026}, while behavioral states remain relatively stable within a local positive-pair window~\citep{markowitz2023reinforcement,weinreb2026tasks}. \Cref{fig:positive_range} (\textit{left}) quantifies the effect of window size. Expanding the positive-pair window beyond the default 10\% of a recording partition increases Day and Proficiency MSE, while Identity accuracy remains relatively stable. This favors local rather than recording-wide temporal neighborhoods for learning-related readouts; the corresponding duration depends on the recording partition (\cref{app:primary_implementation}).

  \mypara{\concepttag{analysisHistory}{How long must a trajectory be to capture identity- and learning-related signals?}}\par\nobreak
  \vspace{-5pt}
  \noindent We examine the effect of trajectory length. As shown in \cref{fig:hist_len} (\textit{right}), longer segments generally improve identity and proficiency prediction by providing more movement context. Interestingly, \ours remains competitive with the default three-second setting even with just one second of input, suggesting that brief movements already contain informative signals, particularly about individual identity.

  \mypara{\concepttag{analysisMovement}{Which movement dynamics reveal identity and learning progress?}} Movement can be examined in egocentric or allocentric reference frames. Here, we decompose \textit{Full Movement} into \textit{Local Pose} and \textit{Global Trajectory}: the former centers and heading-aligns the pose at each frame, while the latter tracks body position and heading over time
  (\cref{fig:traj_view}; \cref{app:traject_view}).
  This allows us to examine how individual identity and learning-related changes are reflected in body posture and movement through space. As shown in~\cref{tab:traj_view}, the two components exhibit different strengths: \textit{Local Pose} is more predictive of identity, whereas \textit{Global Trajectory} is more predictive of learning-related readouts. Combining both yields the strongest overall performance, suggesting that posture dynamics and global movement patterns carry complementary behavioral information.

  \vspace{.5em}
  \begingroup
  \clearcaptionsetup{figure}
  \begin{figure*}[htbp]
    \centering
    \begin{adjustbox}{max width=\linewidth}
      \begin{minipage}{\linewidth}

        \begin{minipage}[t]{0.48\linewidth}
          \vspace{0pt}
          \centering
          \includegraphics[width=\linewidth]{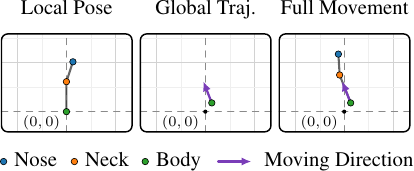}
          \captionsetup{position=bottom,font={rm,small},labelfont=rm,skip=6pt}
          \caption{Decomposition of movement dynamics. Local Pose captures body posture changes, while Global Traj. captures spatial translocation.}
          \label{fig:traj_view}
        \end{minipage}\hfill
        \begin{minipage}[t]{0.49\linewidth}
          \vspace{0pt}
          \centering
          \small
          \captionsetup{type=table,font={rm,small},labelfont=rm,skip=6pt}
          \caption{Different movement dynamics carry complementary behavioral information. Local body dynamics better reveal individuality, while global trajectories better capture learning-related changes.}
          \label{tab:traj_view}
          \setlength{\tabcolsep}{3pt}
          \renewcommand{\arraystretch}{1.12}
          \begin{tabular*}{\linewidth}{@{\extracolsep{\fill}}lccc@{}}
            \toprule
            & \textbf{Identity}$\uparrow$ & \textbf{Day}$\downarrow$ & \textbf{Proficiency}$\downarrow$ \\
            \midrule
            Local Pose & \underline{77.1} & 1.60 & .019 \\
            Global Traj. & 72.1 & \underline{1.47} & \underline{.017} \\
            \midrule
            Full Movement & \textbf{84.0} & \textbf{1.26} & \textbf{.016} \\
            \bottomrule
          \end{tabular*}
        \end{minipage}

      \end{minipage}
    \end{adjustbox}
  \end{figure*}
  \endgroup

  \mypara{\concepttag{analysisIndividuality}{How is individuality expressed in movement?}}

  \begin{wrapfigure}[14]{r}{0.50\textwidth}
    \centering
    \vspace{-20pt}
    \makebox[\linewidth][c]{\includegraphics{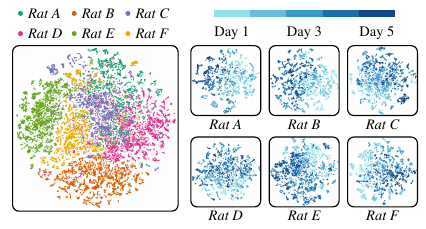}}
    \caption{t-SNE of 3-second trajectories. \textit{Left:} Across animals. \textit{Right:} Across days per rat.}
    \label{fig:tsne_rat}
    \vspace{1em}
  \end{wrapfigure}

  \vspace{-5pt}
  To investigate whether \ours extracts latent representations that reveal individual identity, we project the behavioral latents of all animals into a common t-SNE space (\cref{fig:tsne_rat}); the projection shows that each animal occupies a distinct region. We also fit a separate t-SNE projection for each animal using its latents from all five days. These per-animal views suggest day-related variation, although the extent of separation differs across animals and days can overlap. Together, the visualizations provide qualitative evidence of identity structure and across-day changes, complementing the quantitative results.
  \par\WFclear

  \begin{table*}[!b]
    \centering
    \tablestyle
    \caption{Behavior analysis across maze regions, recording days, and movement motifs. Day prediction is omitted from the per-day analysis. \textbf{Bold} and \underline{underlined} values indicate the best and second-best results within each subtable, based on unrounded values.}

    \begin{minipage}{\linewidth}
      \scriptsize
      \setlength{\tabcolsep}{3pt}
      \renewcommand{\arraystretch}{1.15}
      \setlength{\mazeRowGap}{\dimexpr\ht\strutbox+\dp\strutbox\relax}
      \setlength{\mazeRowGap}{\arraystretch\mazeRowGap}

      \begin{subtable}[t]{0.35\linewidth}
        \centering
        \caption{Maze Regions.}
        \label{tab:maze_region}
        \begin{tabular*}{\linewidth}{@{\extracolsep{\fill}}lccc@{}}
          \toprule
          & \textbf{Identity}$\uparrow$ & \textbf{Day}$\downarrow$ & \textbf{Proficiency}$\downarrow$ \\
          \midrule
          \noalign{\vskip \dimexpr\mazeRowGap/3\relax}
          Arm segment & 81.7 & 1.59 & .021 \\
          \noalign{\vskip \dimexpr2\mazeRowGap/3\relax}
          Arm endpoint & \underline{83.0} & \underline{1.20} & \underline{.015} \\
          \noalign{\vskip \dimexpr2\mazeRowGap/3\relax}
          Junction & \textbf{88.7} & \textbf{1.20} & \textbf{.014} \\
          \noalign{\vskip \dimexpr\mazeRowGap/3\relax}
          \bottomrule
        \end{tabular*}
      \end{subtable}
      \hfill
      \begin{subtable}[t]{0.25\linewidth}
        \centering
        \caption{Recording Days.}
        \label{tab:performance_per_day}
        \begin{tabular*}{\linewidth}{@{\extracolsep{\fill}}lcc@{}}
          \toprule
          & \textbf{Identity}$\uparrow$ & \textbf{Proficiency}$\downarrow$ \\
          \midrule
          Day 1 & 79.3 & .019 \\
          Day 2 & 81.9 & .018 \\
          Day 3 & 78.8 & .016 \\
          Day 4 & \textbf{89.5} & \underline{.013} \\
          Day 5 & \underline{87.6} & \textbf{.012} \\
          \bottomrule
        \end{tabular*}
      \end{subtable}
      \hfill
      \begin{subtable}[t]{0.37\linewidth}
        \centering
        \caption{Movement Motifs.}
        \label{tab:per_motif}
        \begin{tabular*}{\linewidth}{@{\extracolsep{\fill}}lccc@{}}
          \toprule
          & \textbf{Identity}$\uparrow$ & \textbf{Day}$\downarrow$ & \textbf{Proficiency}$\downarrow$ \\
          \midrule
          Remaining still & \underline{87.1} & \underline{1.11} & \underline{.012} \\
          Scanning & 76.6 & 1.34 & .020 \\
          Going straight & 84.0 & 1.37 & .018 \\
          Turning & \textbf{98.0} & \textbf{1.07} & \textbf{.008} \\
          Turning back & 84.2 & 1.56 & .020 \\
          \bottomrule
        \end{tabular*}
      \end{subtable}

    \end{minipage}
  \end{table*}

  \mypara{\concepttag{analysisBehavior}{Under what conditions are identity and learning-related signals most apparent?}}\par\nobreak
  \vspace{-5pt}
  \noindent Having shown that movements contain signatures of identity and learning-related changes, we next ask \textit{whether these signals are expressed continuously or intermittently in the animal's behavior.} We systematically quantify when, where, and during which movements these signals are most apparent (\cref{app:region_motif}). \Cref{tab:maze_region,tab:performance_per_day,tab:per_motif} summarize the results. These signals are strongest at the \textit{maze junction} and clearer on Days 4--5 than on Days 1--3, suggesting that they become more apparent with experience. \textit{Turning} is the most informative motif, consistent with its occurrence primarily at the maze junction, while \textit{remaining still} retains informative signals despite minimal movement.

  \section{Interpretation}

  \subsection{Goal}

  Learning is accompanied by behavior changes that vary across individuals. Our preceding results show that the learned representations carry information about learning progress and individual identity. Here, we explore whether translating these representations back into movement can make these associations visible, offering a qualitative view of the behavioral differences they capture.

  \subsection{Method}
  To connect variation in representation space to visible changes in behavior, we first make the information encoded in these representations observable through generated movement, then vary the representations along directions associated with learning progression and individual differences.

  \mypara{Making representation observable.}
  We train a conditional diffusion model $G_\psi(u,h)$ to generate movement trajectories from a diffusion latent noise $u$ and a conditioning representation $h$ produced by the frozen movement encoder~\citep{peebles2023scalable}. Given an observed trajectory $x_o$ with representation $h_o$, we obtain its latent noise through approximate DDIM inversion.
  We then hold $u_o$ and the model parameters $\psi$ fixed while varying the conditioning representation. The unmodified condition yields the reconstruction $\widehat{x}_o=G_\psi(u_o,h_o)$, providing a baseline for inspecting movement changes induced by representation transformation.

  \mypara{Tracing directions of progression.}
  To trace how movement evolves as learning unfolds, we seek directions that organize progression across recording days, alongside variation across individuals. We estimate these directions using the leading principal components of daily representation centers within an animal or animal-specific centers on the same relative recording day. This provides a shared axis along which to follow gradual changes in movement. Given a selected unit direction $v$, we use an observed representation $h_o$ and a reference representation $h_r$ to define the transformation:
  \begin{equation}
    \label{eq:representation_path}
    \begin{aligned}
      d &= \bigl[(h_r-h_o)^\top v\bigr]v,\\
      h(\tau) &= h_o+\tau d,\qquad \tau\in[0,1],\\
      \widehat{x}(\tau) &= G_\psi\bigl(u_o,h(\tau)\bigr).
    \end{aligned}
  \end{equation}
  Increasing $\tau$ traces a continuous path along this axis toward the reference. Decoding this path with the fixed latent $u_o$ helps preserve the broad motion structure of $x_o$, such as the overall direction of travel, while allowing finer movement patterns to change. This lets us inspect how movement progressively shifts toward characteristics associated with a later recording day or another individual. Further implementation and training details are described in \cref{app:interpretation_directions,app:interpretation_implementation}.

  \resultfigure[!t]
    {\includegraphics[width=397.47253pt,height=182.7227pt]{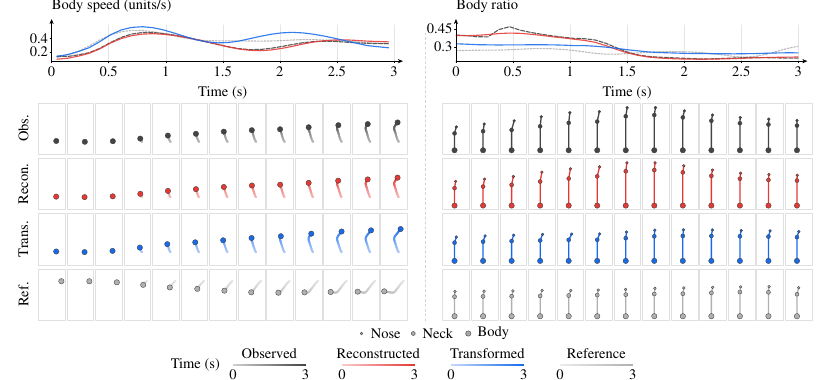}}
    {Turning movement transformations across days and individuals. \textit{Left:} Cross-day Body Trajectory with body speed, the magnitude of body-keypoint velocity. \textit{Right:} Cross-individual Local Pose with body ratio, the nose--neck distance divided by the neck--body distance. Obs., Recon., Trans., and Ref. denote observed, reconstructed, transformed, and reference movements, respectively.}
    {fig:interpretation_progression_turning}

  \subsection{Results}

  We examine movement changes from two complementary perspectives: progression across recording days within an animal, and variation between animals on the same relative recording day. Here, we focus on turning, which carries the strongest day and identity signals, as shown in \cref{tab:per_motif}, making it an informative setting for both comparisons. Both comparisons generate trajectories of all three keypoints. Our earlier findings in~\cref{sec:scientific-analysis} (\cref{tab:traj_view}) show that Global Trajectory predicts day and proficiency more accurately than Local Pose, whereas Local Pose better predicts identity. These findings guide how we inspect the generated movements: through Body Trajectory to follow travel across days, and through Local Pose to follow relative body extension across individuals.

  Across days, we examine how movement execution varies toward a later-day reference. The Body Trajectory sequence on \cref{fig:interpretation_progression_turning} (\textit{left}) traces the body keypoint's path through space, allowing us to follow both where the animal moves and how the turn unfolds. The transformed movement continues in the observed movement's overall direction while taking on aspects of the later-day reference's pace and turning rhythm. The body-speed curve above the sequence shows a corresponding shift toward the reference's speed over time. Together, these observations suggest that the transformation expresses day-related differences in movement execution while preserving the observed movement's broad spatial structure.

  Across individuals, we ask how a movement takes on another animal's postural characteristics. The Local Pose sequence on \cref{fig:interpretation_progression_turning} (\textit{right}) shows how the head, neck, and body extend relative to one another during the turn. As the representation shifts toward the reference animal, the transformed sequence adopts aspects of that animal's pattern of extension, accompanied by a body-ratio curve that moves toward the reference. Together, these observations make individual differences visible in how the head, neck, and body extend relative to one another during the same behavior, connecting the identity prediction results in \cref{tab:traj_view} to visible local postural dynamics. The complementary views---cross-day Local Pose and cross-individual Body Trajectory---are provided in \cref{app:interpretation_complementary}, while \cref{app:interpretation_straight} extends both comparisons to the going-straight motif.

  \section{Discussion and Conclusion}

  \looseness=-1
  In this paper, we present \ours, a framework for learning representations from short-term animal movements. We show that movement ``fingerprints'' can reliably predict individual identity and learning-related readouts. These results suggest that fine-scale movement patterns in freely behaving animals differ across individuals and change systematically over days of learning.

  \looseness=-1
  Beyond predictive performance, \ours provides a tool for generating scientific hypotheses. Through systematic, controlled analyses, we examine when, where, and in which movement patterns identity- and learning-related signals are most apparent. We find that their strength varies across recording days, spatial locations, and movement motifs, and that local pose dynamics and global motion carry complementary information. Conditional generation further allows us to inspect movement differences associated with identity and learning progress, providing hypotheses for subsequent experimental validation.

  \looseness=-1
  Moving forward, we aim to extend \ours to behavioral time-series data from freely moving animals and humans, particularly in experiments where movement patterns evolve over time. By helping researchers identify which aspects of behavior change and how they change, this framework could support more detailed studies of learning. Applied to larger populations, it could also help reveal how learning dynamics vary across individuals.

  \begin{samepage}
  \section{Acknowledgment}
  This research is supported in part by a Whitehall Foundation Research Grant (JYY), a Brain Research Foundation Seed Grant (JYY), and a University of Chicago Artificial Intelligence Initiative seed grant (JYY, ZML). We thank the Ohio Supercomputer Center for generously providing computational resources.
  \end{samepage}

  \clearpage
  \makeatletter
  \begin{center}
    \LARGE\scshape
    \@title\\(Appendix)
  \end{center}
  \makeatother
  \vspace{6pt}
  \appendix
  \raggedbottom
  \addtocontents{toc}{\protect\setcounter{tocdepth}{3}}

  \noindent\textbf{\large Contents of the Appendix}\par
  \vspace{0.5em}
  \begingroup
    \setlength{\parskip}{0pt}
    \hypersetup{linkcolor=blue}
    \makeatletter
    \@starttoc{toc}
    \makeatother
  \endgroup
  \clearpage

  \section{Data, Models, and Evaluation}
  \subsection{Spatial Learning Dataset Statistics}
  \label{app:maze_dataset_statistics}

  \Cref{tab:maze_dataset_statistics} gives per-animal counts for the cohorts in \cref{sec:experiments}. Recording days are indexed from the start of each animal's schedule; later recording phases are excluded.

  Frame counts are taken from the processed trajectory files. Each session is counted once across its training, validation, and test partitions. Recording duration is computed from the first-to-last timestamp span of each session and summed across sessions, excluding intervals between recordings.

  \begin{table}[htbp]
    \centering
    \tablestyle
    \caption{Statistics for the two reward datasets. Frame counts include all data partitions; durations are computed from session timestamps.}
    \label{tab:maze_dataset_statistics}
    \begin{tabular}{@{}lccrr@{}}
      \toprule
      \textbf{Identity} & \textbf{Animal-days} & \textbf{Sessions} & \textbf{Frames} & \textbf{Duration (min)} \\
      \midrule
      \rowcolor[HTML]{F2F5F7}
      \multicolumn{5}{c}{\rule{0pt}{2.3ex}\textit{Alternation Reward}} \\
      \midrule
      \textit{Rat A} & 5 & 15 & 98,708 & 148.9 \\
      \textit{Rat B} & 5 & 20 & 135,646 & 205.2 \\
      \textit{Rat C} & 5 & 20 & 131,844 & 198.9 \\
      \textit{Rat D} & 5 & 20 & 127,143 & 191.8 \\
      \textit{Rat E} & 5 & 20 & 133,732 & 202.2 \\
      \textit{Rat F} & 5 & 20 & 128,259 & 193.5 \\
      \addlinespace[2pt]
      \textbf{Subtotal (6 rats)} & \textbf{30} & \textbf{115} & \textbf{755,332} & \textbf{1,140.5} \\
      \midrule
      \rowcolor[HTML]{F2F5F7}
      \multicolumn{5}{c}{\rule{0pt}{2.3ex}\textit{Probabilistic Reward}} \\
      \midrule
      \textit{Rat G} & 5 & 20 & 130,300 & 196.6 \\
      \textit{Rat H} & 5 & 20 & 132,163 & 199.4 \\
      \textit{Rat I} & 5 & 20 & 130,566 & 197.3 \\
      \textit{Rat J} & 5 & 20 & 92,583 & 139.7 \\
      \textit{Rat K} & 5 & 20 & 135,531 & 204.9 \\
      \textit{Rat L} & 5 & 20 & 135,655 & 205.1 \\
      \textit{Rat M} & 5 & 20 & 136,430 & 206.3 \\
      \textit{Rat N} & 5 & 20 & 135,964 & 205.6 \\
      \addlinespace[2pt]
      \textbf{Subtotal (8 rats)} & \textbf{40} & \textbf{160} & \textbf{1,029,192} & \textbf{1,554.8} \\
      \midrule
      \rowcolor[HTML]{E3E9EE}
      \rule{0pt}{2.5ex}\textbf{Total (14 rats)} & \textbf{70} & \textbf{275} & \textbf{1,784,524} & \textbf{2,695.3} \\
      \bottomrule
    \end{tabular}
  \end{table}

  \Cref{tab:maze_dataset_splits} gives the processed-frame counts for the chronological partitions described in \cref{sec:experiments}.

  \begin{table}[htbp]
    \centering
    \tablestyle
    \caption{Frame counts by partition across two datasets. Rounding at session boundaries produces small deviations from the nominal split proportions.}
    \label{tab:maze_dataset_splits}
    \begin{tabular}{@{}lrrr@{}}
      \toprule
      \textbf{Reward rule} & \textbf{Train} & \textbf{Validation} & \textbf{Test} \\
      \midrule
      \textit{Alternation Reward} & 604,223 & 37,752 & 113,357 \\
      \textit{Probabilistic Reward} & 823,294 & 51,436 & 154,462 \\
      \midrule
      Total & 1,427,517 & 89,188 & 267,819 \\
      \bottomrule
    \end{tabular}
  \end{table}

  \subsection{Implementation Details}
  \label{app:implementation_detail}
  For each method, we selected the configuration with the highest mean of two standardized validation scores on the alternation cohort (6 rats): z-scored Identity accuracy and sign-reversed z-scored Day MSE.

  Classical baselines were fitted using their respective procedures: expectation-maximization for LG-SSM and AR-HMM, Gibbs sampling for Keypoint-MoSeq, and closed-form least squares for ProMPs. The deep baselines (Autoencoder, VAME, T-BYOL, TCN, BAMS, and hBehaveMAE) were trained for 200 epochs, with learning rates selected through validation grid search.

  For LoRA fine-tuning of Chronos-2 and TimesFM-3, we train low-rank adapters for 10 epochs using the self-supervised objective of \ours. The validation criterion above selects the learning rate and LoRA rank.

  \subsubsection{Behavioral Target Construction}
  \label{app:proficiency_computation}

  \mypara{Trial-level proficiency.}
  The supplied state-space labels follow \citet{smith2004dynamic}: a Gaussian random walk models the latent learning state, and a logistic observation model links it to the binary trial outcomes. We use the posterior modes of response probability as trial-level labels.

  We retain entries for all trials, joining each posterior mode to its trial by animal, recording date, session, maze, and trial identifier. Per-day estimates are used where available; trial keys missing from the probabilistic cohort's per-day table use the supplied estimate fitted across recording days. These are retrospective behavioral labels: state-space smoothing uses outcomes later in the fitted sequence as well as earlier ones.

  Let $a_j$ be the start time of trial $j$ and $\hat p_j$ its supplied posterior mode. Between consecutive trial starts, every frame inherits the preceding trial's estimate:
  \begin{equation}
    P(t)=\hat p_j\quad\text{for }a_j\leq t<a_{j+1}.
  \end{equation}
  The first and last trial estimates extend to the recording boundaries. An unmatched trial produces a missing value throughout its interval. For the sampled frame times $\mathcal{T}_i$ covered by history $i$, the probe target is
  \begin{equation}
    y_i^{\mathrm{prof}}=\frac{1}{|\mathcal{T}_i|}\sum_{t\in\mathcal{T}_i}P(t).
  \end{equation}
  A window containing a missing proficiency value is excluded from that target's probe fit and error calculation. Its other targets remain available.

  \mypara{Reward and success rate.}
  Let $m_j$ be the midpoint of trial $j$ and $b_j\in\{0,1\}$ its reward outcome. At frame time $t$, we count all trials with $m_j\in[t-30\,\mathrm{s},t+30\,\mathrm{s})$. Reward is the number of rewarded trials in this interval; Success Rate divides that count by the total number of trials and is zero for an empty interval. Each history receives the mean of its frame-level values. Counts use the complete recording's trial table, including trials across a trajectory-split boundary. These centered targets describe the reward context around the movement.

  \subsubsection{Primary Model and Linear Probes}
  \label{app:primary_implementation}

  \mypara{Preprocessing and windows.}
  We use the registered maze coordinates and downsampled keypoint trajectories from the dataset preprocessing pipeline. Frames with nonfinite tracked coordinates are removed before downsampling. Each recording is split chronologically before constructing stride-one windows, so a trajectory never crosses a partition boundary. The primary feature cache requires 33 history frames and 33 following frames for window eligibility; the encoder receives only the history. After the final-frame alignment described in \cref{app:architecture_detail}, we apply no random rotation, reflection, scaling, or coordinate noise.

  \mypara{Encoder and optimization.}
  The encoder and decoder are trained from scratch. The encoder uses eight attention heads per layer; the reconstruction decoder has four layers with four attention heads each. Both stacks use pre-layer normalization, feed-forward width 256, and dropout 0.1. We train for 200 epochs with AdamW, batch size 256, learning rate $10^{-3}$, $(\beta_1,\beta_2)=(0.9,0.999)$, weight decay 0.01, and gradient-norm clipping at 1. Cosine decay reduces the learning rate to $3\times10^{-5}$. The reported primary run uses seed 42, and features are extracted from the final epoch.

  \mypara{Positive and negative sampling.}
  The sampler uses the history endpoint relative to the beginning of its recording partition. If $D_s$ is that partition's duration and $u_i=t_i^{\mathrm{end}}/D_s$, a positive partner is drawn uniformly from the other eligible windows in the same partition satisfying $|u_i-u_j|\leq0.05$. At an endpoint, the interval is truncated to the available windows; if it contains no other window, the nearest one is used. Thus the nominal neighborhood spans 10\% of the partition. For the 80\% training portion of a ten-minute recording, this is approximately 48 seconds wide, or at most 24 seconds between paired endpoints away from partition boundaries; its duration scales with the partition length. Each epoch shuffles the eligible training windows; incomplete final batches are dropped. Each batch contains 256 anchors and one separately sampled positive for each anchor. For an anchor, the contrastive candidates are its sampled positive and the other batch anchors. Other anchors satisfying the same locality rule are also positives; the remaining candidates are negatives. The anchor itself is excluded. Positive and negative candidate sets receive equal total weight in the balanced loss. Pair construction uses recording membership and time only.

  \mypara{Contrastive objective.}
  We use the balanced objective in \cref{eq:contrastive}, averaged over batch anchors. The implementation assigns total weight $1/2$ to each candidate set; this common factor cancels in the ratio. The numerator normalization contributes an additive $\log|\mathcal{P}|$ term, which is constant with respect to model parameters for fixed candidate sets.

  \mypara{Supervised baseline.}
  The five task losses are summed with unit coefficients, and regression targets retain their original units. The trained encoder is evaluated with the probes below.

  \mypara{Probe fitting.}
  Identity is decoded by one pooled logistic-regression probe across animals, with an $\ell_2$ penalty, $C=1$, and at most 300 L-BFGS iterations. Each continuous target has a separate ridge probe for each animal. Feature means and standard deviations are fitted on the corresponding training data. Ridge penalties are selected by leave-one-out cross-validation on the training rows from $\{10^{-3},10^{-2.5},\ldots,10^3\}$. Encoder configurations are selected using held-out validation scores. Proficiency and Success Rate predictions are clipped to $[0,\,1]$; Day and Reward predictions use the observed target range of the evaluated dataset, pooled across its partitions. Regression MSE is pooled over all valid test windows, weighting each animal by its window count.

  \subsubsection{Hyperparameter Search}

  \Cref{tab:primary_search} lists the hyperparameters considered during model selection. The TRAIL search crosses three learning rates with five reconstruction/contrastive weight pairs, giving 15 configurations; the selected weights are $(1,1)$. Encoder depth is fixed at four layers in this search. The separate architecture sweeps examine latent dimensions $\{1,4,8,16,32,40,48,56,64\}$, history lengths from 1 to 10 seconds, and positive neighborhoods spanning $\{0.5,1,2,5,10,25,50,75,100\}\%$ of a recording partition. The architecture sweeps are reported in the ablation experiments.

  \begin{table}[!htbp]
    \centering
    \small
    \setlength{\tabcolsep}{5pt}
    \renewcommand{\arraystretch}{1.13}
    \caption{Hyperparameter search. Candidate values are grouped by model. Loss weights are listed as (reconstruction, contrastive); Keypoint-MoSeq iteration counts are listed as (AR, full model). Hidden width and HoA weight are fixed at the values shown.}
    \label{tab:primary_search}
    \begin{tabularx}{\linewidth}{@{}>{\raggedright\arraybackslash}p{0.20\linewidth}>{\raggedright\arraybackslash}p{0.25\linewidth}>{\raggedright\arraybackslash}X@{}}
      \toprule
      \textbf{Model} & \textbf{Hyperparameter} & \textbf{Candidate values} \\
      \midrule
      \rowcolor[HTML]{E3ECF5}
      \ours & Learning rate & $10^{-4}$, $3\times10^{-4}$, $10^{-3}$ \\
      \rowcolor[HTML]{E3ECF5}
      & Loss weights & $(1,1)$, $(0.5,1)$, $(1,0.5)$, $(0.25,0.75)$, $(0.75,0.25)$ \\
      \midrule
      \rowcolor[HTML]{F2F5F7}
      \multicolumn{3}{l}{\textit{Classic methods}} \\
      \addlinespace[2pt]
      LG-SSM & Latent dimension & 4, 8, 16, 32, 64, 128 \\
      & EM iterations & 100, 200 \\
      \addlinespace[3pt]
      AR-HMM & Number of states & 32, 64, 128 \\
      & Autoregressive lags & 1, 3 \\
      & EM iterations & 100, 200 \\
      \addlinespace[3pt]
      ProMPs & Basis weights & 2, 3, 5, 8, 10, 12, 15, 18, 20, 24, 28, 33, 36, 40, 50 \\
      \addlinespace[3pt]
      Keypoint-MoSeq & Number of states & 20, 30, 50 \\
      & Latent dimension & 6, 10 \\
      & AR / full iterations & $(100,100)$, $(75,125)$, $(50,150)$ \\
      \midrule
      \rowcolor[HTML]{F2F5F7}
      \multicolumn{3}{l}{\textit{Deep learning methods}} \\
      \addlinespace[2pt]
      Autoencoder & Layers & 2, 4, 6, 8 \\
      & Learning rate & $10^{-4}$, $3\times10^{-4}$, $10^{-3}$ \\
      \addlinespace[3pt]
      TCN / T-BYOL & Channels per layer & 16, 32, 64, 128, 256 (three equal-width layers) \\
      & Learning rate & $10^{-4}$, $3\times10^{-4}$, $10^{-3}$ \\
      \addlinespace[3pt]
      VAME & Latent dimension & 64, 128 \\
      & Hidden width & 256 \\
      & $\beta$ & 1, 2 \\
      & Learning rate & $10^{-4}$, $3\times10^{-4}$, $5\times10^{-4}$, $10^{-3}$ \\
      \addlinespace[3pt]
      BAMS & HoA weight & 500 \\
      & HoA bins & 16, 32 \\
      & BYOL weight & 0.5, 1 \\
      & Learning rate & $10^{-4}$, $3\times10^{-4}$, $10^{-3}$ \\
      \midrule
      \rowcolor[HTML]{F2F5F7}
      \multicolumn{3}{l}{\textit{Foundation-model fine-tuning}} \\
      \addlinespace[2pt]
      LoRA & Adapter rank & 4, 8, 16 \\
      & Learning rate & $10^{-5}$, $3\times10^{-5}$, $10^{-4}$, $3\times10^{-4}$, $10^{-3}$ \\
      \bottomrule
    \end{tabularx}
  \end{table}

  \subsection{Encoder Settings for the Additional Experiments}
  \begin{table}[!htbp]
    \centering
    \tablestyle
    \caption{Encoder configurations used in the additional experiments. Both use multi-frequency coordinate features, 64-dimensional tokens, eight queries, four encoder layers, and contrastive temperature 0.1. The performance rows use the same 105{,}882 test windows and state-space Proficiency labels. Identity is percentage accuracy; the remaining targets are MSE.}
    \label{tab:additional_configurations}
    \begin{tabularx}{\linewidth}{lXX}
      \toprule
      \textbf{Setting} & \textbf{Primary} & \textbf{Supplementary} \\
      \midrule
      Coordinate projection & Linear & Two-layer MLP \\
      Maze tokens & Waypoints & Vertices and edge descriptors \\
      First-query key normalization & None & Per-head layer normalization \\
      Reconstruction input & Restored representation and seven queries & Full encoder query bank \\
      Reconstruction coefficient & 1 & 0.5 \\
      \midrule
      Identity & 84.00 & 84.07 \\
      Day & 1.258 & 1.252 \\
      Proficiency & .01556 & .01552 \\
      Reward & 0.609 & 0.593 \\
      \bottomrule
    \end{tabularx}
  \end{table}

  Within-session experiments use the primary data split. Session- and day-held-out experiments keep complete recordings together, as specified in \cref{app:additional_analysis}. We freeze the encoder before fitting probes, standardize features using the training partition, and select ridge penalties from $\{10^{-3},10^{-2.5},\ldots,10^3\}$ on the validation partition. Identity uses regularized logistic regression. Proficiency, rewarded count, and success rate follow the target definitions in \cref{app:proficiency_computation}.

  \subsection{Uncertainty Estimates}

  We quantify uncertainty by resampling animals, keeping all test windows from a selected animal together. For the primary cohort, movement-view, and reconstruction comparisons, we recompute pooled accuracy or MSE from the selected animals' window counts and score sums. The generalization tables use equal-animal averages, as indicated in their captions. The 2.5th and 97.5th percentiles give the reported 95\% confidence intervals. For the primary six-animal alternation estimates, we enumerate all $6^6$ bootstrap resamples. For the eight-animal probabilistic cohort, we draw 20,000 resamples with random seed 42. The encoder and probes remain fixed, so these intervals describe variation across animals under the fitted models. For paired comparisons, each bootstrap draw uses the same animals for both methods.

  \FloatBarrier
  \section{Prediction Across Datasets and Partitions}
  \subsection{Performance Across Reward Conditions}
  \label{app:alternation_results}
  \Cref{tab:yshape_probability} gives the full comparison for probabilistic rewards. \Cref{tab:primary_cohort_uncertainty} reports animal-bootstrap intervals for TRAIL in both cohorts. The paired comparisons in \cref{fig:primary_paired_comparisons} use the primary alternation runs from \cref{tab:yshape_alternation}.

  \begin{table}[!h]
    \centering
    \tablestyle
    \caption{Performance on the spatial-learning dataset (probabilistic rewards).}
    \label{tab:yshape_probability}
    \begin{adjustbox}{max width=\linewidth}
      \begin{tabular}{llccccc}
        \toprule
        & \multicolumn{1}{c}{\multirow{2}{*}{\textbf{Method}}} & \textbf{Identity} & \textbf{Day} & \textbf{Proficiency} & \textbf{Success Rate} & \textbf{Reward} \\
        & & (Acc $\uparrow$) & (MSE $\downarrow$) & (MSE $\downarrow$) & (MSE $\downarrow$) & (MSE $\downarrow$) \\
        \midrule
        \multirow{5}{*}{\textbf{Classic}} & Direct Trajectory & 35.18 & 1.89 & .011 & 0.07 & 0.91 \\
        & LG-SSM\phantom{~\citep{Durbin_2012}} & 19.40 & 1.88 & .011 & 0.05 & 0.85 \\
        & AR-HMM\phantom{~\citep{Wiltschko_2015}} & 35.40 & 1.72 & .010 & 0.07 & 0.83 \\
        & ProMPs\phantom{~\citep{NIPS2013_e53a0a29}} & 29.96 & 1.84 & .010 & 0.05 & 0.83 \\
        & Keypoint-MoSeq\phantom{~\citep{Weinreb2024}} & 33.58 & 1.94 & .011 & 0.07 & 0.94 \\
        \midrule
        \multirow{4}{*}{\textbf{Foundation Model}} & TimesFM-3 (Zero-shot)\phantom{~\citep{google2026timesfm3}} & 31.73 & 1.87 & .011 & 0.08 & 0.91 \\
        & TimesFM-3 (LoRA) & 36.42 & 1.82 & .010 & 0.08 & 0.89 \\
        & Chronos-2 (Zero-shot)\phantom{~\citep{ansari2025chronos2}} & 33.33 & 1.81 & .010 & 0.08 & 0.89 \\
        & Chronos-2 (LoRA) & 41.06 & 1.73 & .010 & 0.08 & 0.86 \\
        \midrule
        \multirow{8}{*}{\textbf{Deep Learning}} & AutoEncoder\phantom{~\citep{Hinton_2006}} & 55.63 & 1.67 & \underline{.009} & \underline{0.05} & 0.73 \\
        & VAME\phantom{~\citep{Luxem2022}} & 59.49 & 1.72 & .009 & 0.05 & 0.75 \\
        & T-BYOL\phantom{~\citep{NEURIPS2020_f3ada80d}} & 56.72 & 1.78 & .010 & 0.05 & 0.79 \\
        & TCN\phantom{~\citep{BaiTCN2018}} & 66.10 & \underline{1.61} & .009 & 0.05 & 0.81 \\
        & BAMS\phantom{~\citep{NEURIPS2023_5aad86aa}} & 62.40 & 1.72 & .009 & 0.05 & 0.73 \\
        & hBehaveMAE\phantom{~\citep{Stoffl2025hBehaveMAE}} & 46.58 & 1.73 & .009 & \textbf{0.05} & \underline{0.71} \\
        & Supervised & \underline{82.26} & 1.75 & .009 & 0.06 & 0.91 \\
        \rowcolor[HTML]{E3ECF5}       \cellcolor{white} & \ours & \textbf{85.87} & \textbf{1.33} & \textbf{.007} & 0.05 & \textbf{0.68} \\
        \bottomrule
      \end{tabular}
    \end{adjustbox}
  \end{table}

  \begin{table}[!htbp]
    \centering
    \tablestyle
    \caption{TRAIL prediction across reward conditions. Estimates use the held-out predictions from the cohort comparisons; confidence intervals are obtained by bootstrapping animals at the 95\% level. Identity is percentage accuracy; the remaining targets are MSE.}
    \label{tab:primary_cohort_uncertainty}
    \begin{tabular}{llrr}
      \toprule
      \textbf{Cohort} & \textbf{Target} & \textbf{Estimate} & \textbf{Confidence interval} \\
      \midrule
      Alternation & Identity (\%) & 84.00 & $[76.24,\,91.10]$ \\
      Alternation & Day & 1.2579 & $[0.9569,\,1.5937]$ \\
      Alternation & Proficiency & .0156 & $[.0109,\,.0223]$ \\
      Alternation & Success Rate & 0.13 & $[0.09,\,0.18]$ \\
      Alternation & Reward & 0.6090 & $[0.4488,\,0.8493]$ \\
      \midrule
      Probabilistic & Identity (\%) & 85.87 & $[82.59,\,88.81]$ \\
      Probabilistic & Day & 1.3329 & $[1.0275,\,1.6831]$ \\
      Probabilistic & Proficiency & .0071 & $[.0054,\,.0087]$ \\
      Probabilistic & Success Rate & 0.05 & $[0.04,\,0.06]$ \\
      Probabilistic & Reward & 0.6824 & $[0.6094,\,0.7657]$ \\
      \bottomrule
    \end{tabular}
  \end{table}

  \resultfigure
    {\includegraphics{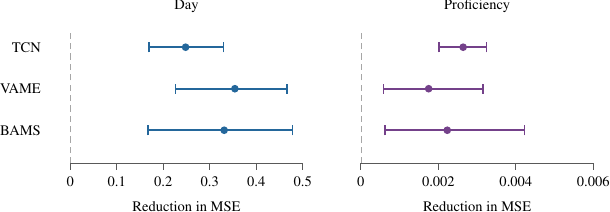}}
    {Paired comparisons in the primary alternation cohort. Points show pooled MSE reductions relative to each baseline; horizontal bars show animal-bootstrap 95\% confidence intervals. All methods use the same 105,882 test windows.}
    {fig:primary_paired_comparisons}

  \subsection{Long-Context Prediction}
  \Cref{tab:yshape_long_context} compares \ours with BAMS and hBehaveMAE in the long-context setting. BAMS uses both 3\,s (33-frame) short-term and 10\,s (110-frame) long-term windows, while \ours and hBehaveMAE take 10\,s trajectories as input.

  \begin{table}[htpb]
    \centering
    \tablestyle
    \caption{Long-context prediction in both reward groups. \tableranking}
    \label{tab:yshape_long_context}
    \begin{adjustbox}{max width=\linewidth}
      \begin{tabular}{lccccc}
        \toprule
        \textbf{Method} & \textbf{Identity} ($\uparrow$) & \textbf{Day} ($\downarrow$) & \textbf{Proficiency} ($\downarrow$) & \textbf{Success Rate} ($\downarrow$) & \textbf{Reward} ($\downarrow$) \\
        \midrule
        \rowcolor[HTML]{F2F5F7}
        \multicolumn{6}{c}{\rule{0pt}{2.3ex}\textit{Alternation Reward}} \\
        \midrule
        BAMS & \underline{73.68} & \underline{1.32} & .017 & \textbf{0.12} & \underline{0.54} \\
        hBehaveMAE & 50.87 & 1.55 & \underline{.017} & \textbf{0.12} & 0.59 \\
        \rowcolor[HTML]{F2F5F7}
        \ours & \textbf{85.26} & \textbf{1.10} & \textbf{.014} & \underline{0.13} & \textbf{0.51} \\
        \midrule
        \rowcolor[HTML]{F2F5F7}
        \multicolumn{6}{c}{\rule{0pt}{2.3ex}\textit{Probabilistic Reward}} \\
        \midrule
        BAMS & \underline{77.64} & \underline{1.52} & \underline{.008} & \textbf{0.04} & \underline{0.62} \\
        hBehaveMAE & 48.65 & 1.68 & .009 & \textbf{0.04} & 0.64 \\
        \rowcolor[HTML]{F2F5F7}
        \ours & \textbf{87.87} & \textbf{1.29} & \textbf{.006} & \underline{0.04} & \textbf{0.58} \\
        \bottomrule
      \end{tabular}
    \end{adjustbox}
  \end{table}

  \subsection{MABe22 Mouse Triplets}
  \label{app:mabe_evaluation}

  The short-context TRAIL model uses 61 frames at the native 30 Hz and produces a 64-dimensional representation for each mouse. For the official tasks, we concatenate the coordinate-wise mean and range of the three mice's representations at each frame, yielding 128 features. The encoder is fitted on \texttt{UserTrain} and frozen before feature extraction. Task probes use all of \texttt{SubmissionTrain}, ridge penalty 1, and seed 42; classification uses class-balanced ridge regression. The reported benchmark scores use the official private test split and its task-specific scoring rules.

  In \cref{tab:mabe22}, Tasks MSE averages the experiment-day and time-of-day regression errors and multiplies the result by 100. Tasks F1 averages the 11 classification scores. Frame F1 averages the nine interaction tasks, and Seq. F1 averages strain and lighting classification. Each F1 column is expressed as a percentage.

  For within-clip Identity, a separate class-balanced ridge classifier predicts the three mouse slots in each test clip. Feature standardization uses that clip's first 42 seconds, and the ridge penalty is 1. We score valid frames in its final 18 seconds and average accuracy equally across clips. \Cref{tab:mabe_identity_uncertainty} compares the short-context TRAIL and BAMS runs on the same 1,058 clips, with identical numbers of valid training and test frames per mouse. Confidence intervals use 20,000 paired clip-bootstrap resamples with seed 42, keeping the encoders and fitted probes fixed.

  \begin{table}[!htbp]
    \centering
    \tablestyle
    \caption{Within-clip identity on MABe22 Mouse Triplets. Estimates average the accuracies of the 1,058 test clips. The 95\% confidence intervals use paired clip-bootstrap resampling; mouse slots are defined separately in each clip. The final row reports TRAIL's accuracy improvement over BAMS in percentage points.}
    \label{tab:mabe_identity_uncertainty}
    \begin{tabular}{lrr}
      \toprule
      \textbf{Representation} & \textbf{Accuracy (\%)} & \textbf{Confidence interval} \\
      \midrule
      BAMS, short context & 63.08 & $[61.72,\,64.46]$ \\
      \rowcolor[HTML]{E3ECF5}
      \ours, short context & 73.65 & $[72.28,\,75.01]$ \\
      \midrule
      Accuracy improvement (pp) & 10.56 & $[9.68,\,11.45]$ \\
      \bottomrule
    \end{tabular}
  \end{table}

  \subsection{Window Selection and Repeated Training}
  The generalization, window-selection, repeated-training, movement-descriptor, and maze-context experiments use the supplementary encoder in \cref{tab:additional_configurations}. It shares the primary model's trajectory alignment, temporal sampling, contrastive objective, and optimization schedule. Runs use seed 42 unless a table specifies otherwise. The temporal-neighborhood comparison uses the primary model. Both encoders are evaluated below on the same within-session test windows.

  These experiments use the supplementary encoder. We remove windows within 30 seconds of split boundaries and retain stride-33 windows within each session. The frozen-feature probes use identical training and test windows for TRAIL and BAMS. TRAIL retains lower day, proficiency, and rewarded-count MSE in all six animals (\cref{tab:additional_stability}). The same table reports the mean and standard deviation from three training seeds under the dense-window protocol.

  \begin{table}[!htbp]
    \centering
    \tablestyle
    \caption{Window selection and training variability. Identity is accuracy; the remaining readouts are MSE. The nonoverlap comparison uses 17{,}102 training and 2{,}004 test windows. Intervals below the MSE reductions are paired animal-bootstrap 95\% confidence intervals. The final row uses seeds 7, 42, and 2026 with validation-selected checkpoints.}
    \label{tab:additional_stability}
    \begin{adjustbox}{max width=\linewidth}
      \begin{tabular}{lrrrr}
        \toprule
        \textbf{Setting} & \textbf{Identity} & \textbf{Day} & \textbf{Proficiency} & \textbf{Reward} \\
        \midrule
        BAMS, nonoverlap & 64.97\% & 1.847 & .02117 & 1.032 \\
        \rowcolor[HTML]{E3ECF5}
        TRAIL, nonoverlap & 84.98\% & 1.407 & .01634 & 0.789 \\
        MSE reduction & -- & 0.441 & .00478 & 0.242 \\
         & -- & $[0.289,\,0.593]$ & $[.00337,\,.00659]$ & $[0.141,\,0.357]$ \\
        \midrule
        \multicolumn{5}{l}{\rule{0pt}{2.3ex}\textit{Dense-window protocol: mean $\pm$ SD over three seeds}} \\
        TRAIL, three seeds & $83.72\%\pm0.45\%$ & $1.273\pm0.020$ & $.01606\pm.00063$ & $0.607\pm0.013$ \\
        \bottomrule
      \end{tabular}
    \end{adjustbox}
  \end{table}

  \subsection{Generalization Across Sessions, Days, and Animals}
  \label{app:additional_analysis}

  \subsubsection{Held-Out Sessions}
  For each animal and recording day, complete sessions are sorted in acquisition order. Sessions before the final two are used for training, the penultimate session for validation, and the final session for testing. All original temporal shards of a session remain together. This gives 55 training, 30 validation, and 30 test sessions, with all six animals and five recording days represented in the test set.

  We compare the supplementary encoder with BAMS using 26-channel local kinematics and maze geometry in the same body-centered frame. BAMS uses 33-frame and 110-frame histories, its HoA and BYOL objectives, seed 42, and 200 training epochs. Both methods are evaluated at the same endpoints eligible for the longer history. \Cref{tab:additional_session_bams} shows lower error for day, proficiency, and rewarded count.

  \begin{table}[!htbp]
    \centering
    \tablestyle[3pt]
    \caption{Prediction on held-out sessions. MSE is averaged equally across animals. MSE reductions measure the improvement with TRAIL over BAMS, with paired animal-bootstrap 95\% confidence intervals. The comparison uses 183{,}884 windows, except proficiency, which has 181{,}639 complete cases. BAMS uses local coordinates and maze context. The final column counts animals with lower MSE under TRAIL.}
    \label{tab:additional_session_bams}
    \begin{tabular}{@{}lrrrrr@{}}
      \toprule
      \textbf{Target} & TRAIL & BAMS & \textbf{Reduction} & \textbf{Confidence interval} & \textbf{Animals} \\
      \midrule
      Day & 1.765 & 1.932 & $0.167$ & $[0.071,\,0.255]$ & 5 of 6 \\
      Proficiency & .02032 & .02460 & $.00428$ & $[.00124,\,.00897]$ & 5 of 6 \\
      Reward & 0.760 & 1.179 & $0.419$ & $[0.131,\,0.788]$ & 6 of 6 \\
      \bottomrule
    \end{tabular}
  \end{table}

  We also compare the held-out-session encoder with an animal-specific training mean and a ridge probe using 14 kinematic features. The representation gives lower day and proficiency error than either baseline in five of the six animals (\cref{tab:additional_session_readouts}).

  \begin{table}[!htbp]
    \centering
    \tablestyle[3pt]
    \caption{Day and proficiency readouts on held-out sessions. MSE is averaged equally across six animals. The lower block reports the reduction in MSE achieved by TRAIL relative to each baseline, with paired animal-bootstrap 95\% confidence intervals. Day uses 190,814 test windows, and Proficiency uses 188,261 complete cases.}
    \label{tab:additional_session_readouts}
    \begin{tabular}{lrr}
      \toprule
      \textbf{Readout} & \textbf{Day} & \textbf{Proficiency} \\
      \midrule
      Animal-specific mean & 2.012 & .02571 \\
      14 kinematic features & 1.950 & .02541 \\
      \rowcolor[HTML]{E3ECF5}
      \ours & 1.756 & .02046 \\
      \bottomrule
    \end{tabular}

    \medskip
    \begin{tabular}{@{}lrrrr@{}}
      \toprule
      \textbf{Baseline} & \multicolumn{2}{c}{\textbf{Day}} & \multicolumn{2}{c}{\textbf{Proficiency}} \\
      \cmidrule(lr){2-3}\cmidrule(lr){4-5}
      & Reduction & Confidence interval & Reduction & Confidence interval \\
      \midrule
      Training mean & $0.256$ & $[0.051,\,0.430]$ & $.00524$ & $[.00143,\,.00922]$ \\
      Kinematic features & $0.194$ & $[0.054,\,0.316]$ & $.00495$ & $[.00230,\,.00753]$ \\
      \bottomrule
    \end{tabular}
  \end{table}

  \subsubsection{Held-Out Day}
  We train encoders and probes on days 1--3, select probe regularization on day 4, and test on day 5, keeping complete sessions and their temporal shards together. On day 5, TRAIL achieves higher identity accuracy than BAMS and simple kinematics, and lower rewarded-count error than BAMS (\cref{tab:additional_day_identity}).

  \begin{table}[!htbp]
    \centering
    \tablestyle[3pt]
    \caption{Identity and rewarded-count prediction on a held-out day. Days 1--3 train the encoders and probes; day 4 selects regularization. Results use 139{,}875 day-5 windows. TRAIL improves identity accuracy in all six animals; gains are measured against each listed baseline in percentage points. Rewarded-count MSE is averaged equally across animals; five of six animals improve. Intervals are paired animal-bootstrap 95\% confidence intervals.}
    \label{tab:additional_day_identity}
    \begin{tabular}{lrrr}
      \toprule
      \textbf{Representation} & \textbf{Balanced acc. (\%)} & \textbf{Gain (pp)} & \textbf{Confidence interval} \\
      \midrule
      Chance & 16.67\% & 58.21 pp & $[44.08,\,72.33]$ \\
      Kinematic features & 25.90\% & 48.98 pp & $[33.91,\,62.35]$ \\
      BAMS (local + maze) & 38.89\% & 35.99 pp & $[20.80,\,55.76]$ \\
      \rowcolor[HTML]{E3ECF5}
      \ours & \textbf{74.87\%} & -- & -- \\
      \bottomrule
    \end{tabular}

    \medskip
    \begin{tabular}{@{}lrrrr@{}}
      \toprule
      \textbf{Target} & TRAIL MSE & BAMS MSE & \textbf{Reduction} & \textbf{Confidence interval} \\
      \midrule
      Reward & 0.948 & 1.608 & $0.660$ & $[0.015,\,1.790]$ \\
      \bottomrule
    \end{tabular}
  \end{table}

  \subsubsection{Within-Recording Temporal Alignment}
  We keep the day-5 encoders and day-4-selected rewarded-count probes fixed, then permute features within each test shard. The analysis uses 4{,}275 nonoverlapping windows from 69 shards and six animals, with 1{,}000 permutations. Shuffling increases prediction error for both representations (\cref{tab:additional_time_alignment}), indicating that their features track changes in rewarded count within a recording.

  \begin{table}[!htbp]
    \centering
    \tablestyle[3pt]
    \caption{Within-recording time alignment. Errors are mean per-animal rewarded-count MSE. The reduction compares correctly aligned features with permuted features; intervals are animal-bootstrap 95\% confidence intervals. Each method improves in five of six animals (one-sided permutation $p=0.001$).}
    \label{tab:additional_time_alignment}
    \begin{tabular}{@{}lrrrr@{}}
      \toprule
      \textbf{Representation} & \textbf{Aligned} & \textbf{Permuted} & \textbf{Reduction} & \textbf{Confidence interval} \\
      \midrule
      BAMS (local + maze) & 1.593 & 1.661 & $0.068$ & $[0.0031,\,0.1604]$ \\
      \rowcolor[HTML]{E3ECF5}
      \ours & 0.948 & 0.980 & $0.032$ & $[0.0029,\,0.0683]$ \\
      \bottomrule
    \end{tabular}
  \end{table}

  \subsubsection{Transfer to New Animals}
  An encoder trained and selected on ten animals is frozen before evaluation on four different animals. Each new animal supplies a chronological training partition for fitting its own ridge probe; later windows are held out for testing. TRAIL has lower error than the kinematic features on every target in all four animals (\cref{tab:additional_animal_transfer}).

  \begin{table}[!htbp]
    \centering
    \tablestyle[3pt]
    \caption{Transfer with calibration on four new animals. MSE is averaged equally across the four animals using 74{,}768 test windows. MSE reductions measure the improvement with TRAIL over kinematic features, with paired animal-bootstrap 95\% confidence intervals.}
    \label{tab:additional_animal_transfer}
    \begin{tabular}{@{}lrrrr@{}}
      \toprule
      \textbf{Target} & TRAIL & Kinematics & \textbf{Reduction} & \textbf{Confidence interval} \\
      \midrule
      Day & 1.621 & 1.861 & $0.241$ & $[0.182,\,0.284]$ \\
      Proficiency & .00944 & .01112 & $.00167$ & $[.00073,\,.00261]$ \\
      Reward & 0.860 & 1.025 & $0.165$ & $[0.082,\,0.249]$ \\
      Success Rate & 0.05 & 0.07 & $0.02$ & $[0.01,\,0.02]$ \\
      \bottomrule
    \end{tabular}
  \end{table}

  \FloatBarrier
  \section{Model Design and Movement Analysis}
  \subsection{Model Design}
  \label{app:ablation}
  \begin{table*}[htbp]
    \centering
    \tablestyle
    \caption{Ablation studies on \concepttag{ablationRecon}{loss function} and \concepttag{ablationMultiFreq}{multi-frequency modeling}. \textbf{Bold}: best based on unrounded values.}

    \begin{subtable}[t]{0.49\linewidth}
      \centering
      \caption{Reconstruction loss.}
      \label{tab:reconstruction_ablation}
      \small
      \setlength{\tabcolsep}{1.5pt}
      \begin{tabularx}{\linewidth}{l*{2}{>{\centering\arraybackslash}X}}
        \toprule
        \textbf{Target} & Without & With \\
        \midrule
        Identity$\uparrow$ & 81.1 & \cellcolor{ablationRecon}\textbf{84.0} \\
        Day$\downarrow$ & 1.32 & \cellcolor{ablationRecon}\textbf{1.26} \\
        Proficiency$\downarrow$ & .017 & \cellcolor{ablationRecon}\textbf{.016} \\
        Success Rate$\downarrow$ & 0.14 & \cellcolor{ablationRecon}\textbf{0.13} \\
        Reward$\downarrow$ & 0.63 & \cellcolor{ablationRecon}\textbf{0.61} \\
        \bottomrule
      \end{tabularx}
    \end{subtable}
    \hfill
    \begin{subtable}[t]{0.49\linewidth}
      \centering
      \caption{{Multi-frequency modeling}.}
      \label{tab:multiple_frequencies_ablation}
      \small
      \setlength{\tabcolsep}{1.5pt}
      \begin{tabularx}{\linewidth}{l*{2}{>{\centering\arraybackslash}X}}
        \toprule
        \textbf{Target} & Without & With \\
        \midrule
        Identity$\uparrow$ & 83.2 & \cellcolor{ablationMultiFreq}\textbf{84.0} \\
        Day$\downarrow$ & 1.28 & \cellcolor{ablationMultiFreq}\textbf{1.26} \\
        Proficiency$\downarrow$ & .016 & \cellcolor{ablationMultiFreq}\textbf{.016} \\
        Success Rate$\downarrow$ & \textbf{0.13} & \cellcolor{ablationMultiFreq}0.13 \\
        Reward$\downarrow$ & 0.61 & \cellcolor{ablationMultiFreq}\textbf{0.61} \\
        \bottomrule
      \end{tabularx}
    \end{subtable}
  \end{table*}

  \begin{wrapfigure}{r}{0.5\linewidth}
    \centering
    \vspace{-1em}
    \includegraphics{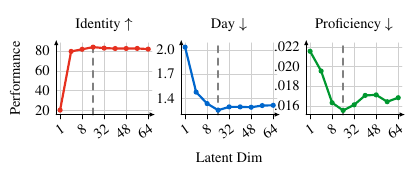}
    \captionsetup{skip=3pt}
    \caption{\concepttag{ablationLatent}{Latent representation} trade-off. The default representation has 16 dimensions.}
    \label{fig:rep_dim}
  \end{wrapfigure}

  \mypara{\concepttag{ablationRecon}{Reconstruction loss.}}
  We analyze \ours with and without reconstruction loss in \cref{tab:reconstruction_ablation}. The model trained with reconstruction has higher identity accuracy and lower day and proficiency errors. \Cref{fig:primary_component_comparisons} reports the reduction in prediction error.

  \mypara{\concepttag{ablationMultiFreq}{Multi-frequency modeling.}}
  We evaluate the contribution of multi-frequency modeling in \cref{tab:multiple_frequencies_ablation}. The multi-frequency features give modest improvements in identity, day, and proficiency prediction in this comparison.

  \mypara{\concepttag{ablationLatent}{Latent representation.}}
  We present the effect of latent dimensionality on identity and learning-related prediction in \cref{fig:rep_dim}. The default uses 16 dimensions; the eight-dimensional representation also performs well across the three readouts.
  \par\WFclear

  \subsubsection{Temporal Neighborhoods}
  Using the primary encoder and the sampler in \cref{app:primary_implementation}, we compare neighborhoods spanning 10\% and 100\% of a recording partition. Both runs use seed 42 and the same test windows. The narrower neighborhood gives lower day and proficiency error in all six animals (\cref{tab:additional_temporal_control}).

  \begin{table}[!htbp]
    \centering
    \tablestyle
    \caption{Effect of positive-neighborhood width. Identity is accuracy; Day and Proficiency are MSE. Results use 105{,}882 windows. The final rows give mean per-animal MSE reductions, with paired animal-bootstrap 95\% confidence intervals directly below.}
    \label{tab:additional_temporal_control}
    \begin{adjustbox}{max width=\linewidth}
      \begin{tabular}{lrrr}
        \toprule
        \textbf{Positive sampling} & \textbf{Identity} & \textbf{Day} & \textbf{Proficiency} \\
        \midrule
        Broad 100\% & 82.37\% & 1.465 & .0211 \\
        \rowcolor[HTML]{E3ECF5}
        Local 10\% & 84.00\% & 1.258 & .0156 \\
        \midrule
        MSE reduction & -- & 0.211 & .00541 \\
         & -- & $[0.096,\,0.370]$ & $[.00144,\,.00947]$ \\
        \bottomrule
      \end{tabular}
    \end{adjustbox}
  \end{table}

  \subsubsection{Reconstruction Inputs}
  On a two-animal subset, we train the decoder for 100 epochs with either all history queries or only the restored representation, using seed 42. \Cref{tab:additional_reconstruction} reports validation performance for both inputs.

  \begin{table}[!htbp]
    \centering
    \tablestyle
    \caption{Reconstruction-context comparison on two animals. Identity is accuracy; the four regression columns report MSE. The last column is reconstruction negative log-likelihood. Results use 5,250 validation windows.}
    \label{tab:additional_reconstruction}
    \begin{adjustbox}{max width=\linewidth}
      \begin{tabular}{lrrrrrr}
        \toprule
        \textbf{Context} & \textbf{Identity} & \textbf{Day} & \textbf{Proficiency} & \textbf{Success Rate} & \textbf{Reward} & \textbf{NLL} \\
        \midrule
        All queries & 87.98\% & 2.0573 & .0008230 & 0.11 & 0.42644 & $-6.731$ \\
        Representation only & 87.09\% & 2.0578 & .0008240 & 0.11 & 0.42615 & $-6.733$ \\
        \bottomrule
      \end{tabular}
    \end{adjustbox}
  \end{table}

  \subsection{Movement Views}
  \label{app:traject_view}

  \textit{Full Movement} uses the final-frame alignment in \cref{app:architecture_detail}.

  For \textit{Local Pose}, we apply this standardization independently to each frame, ensuring that the body keypoint remains at the origin and the body--neck segment remains vertically aligned. This removes global translation and rotation while retaining changes in relative body configuration. For \textit{Global Trajectory}, we retain only the body coordinates from the Full Movement representation for global position, along with the body--neck unit vector as the orientation over time.

  \begin{table}[!htbp]
    \centering
    \tablestyle[3pt]
    \caption{Prediction from local pose and global trajectory. Pooled estimates use the same 105,882 test windows as \cref{tab:traj_view}. Intervals below the estimates are animal-bootstrap 95\% confidence intervals. Identity is percentage accuracy; Day and Proficiency are MSE.}
    \begin{tabular}{@{}lccc@{}}
      \toprule
      \textbf{View} & \textbf{Identity} & \textbf{Day} & \textbf{Proficiency} \\
      \midrule
      Local Pose & 77.1 & 1.6047 & .0189 \\
      & $[66.2,\,87.4]$ & $[1.3303,\,1.9455]$ & $[.0124,\,.0277]$ \\
      \addlinespace[4pt]
      Global Trajectory & 72.1 & 1.4733 & .0172 \\
      & $[61.4,\,82.3]$ & $[1.3063,\,1.6259]$ & $[.0128,\,.0231]$ \\
      \addlinespace[4pt]
      \rowcolor[HTML]{E3ECF5}
      Full Movement & 84.0 & 1.2579 & .0156 \\
      \rowcolor[HTML]{E3ECF5}
      & $[76.2,\,91.1]$ & $[0.9569,\,1.5937]$ & $[.0109,\,.0223]$ \\
      \addlinespace[4pt]
      \bottomrule
    \end{tabular}
  \end{table}

  \Cref{fig:primary_component_comparisons} gives paired error reductions for the movement-view and reconstruction comparisons. Each comparison measures the improvement of the complete primary model over the indicated variant.

  \resultfigure
    {\includegraphics{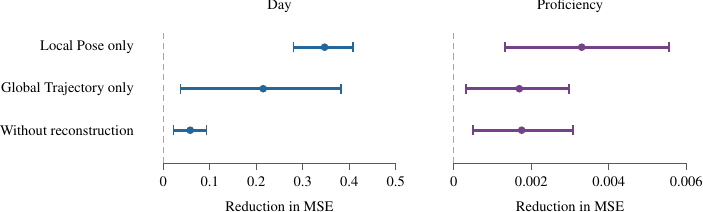}}
    {Movement inputs and reconstruction. Points show pooled MSE reductions relative to each ablated model; horizontal bars show animal-bootstrap 95\% confidence intervals. All methods use the same 105,882 test windows. The comparisons use the runs in \cref{tab:traj_view,tab:reconstruction_ablation}.}
    {fig:primary_component_comparisons}

  \subsection{Variation Across Animals}
  \Cref{tab:performance_per_rat} reports each animal's test performance using the primary encoder and probes.

  \begin{table}[htpb]
    \centering
    \tablestyle
    \caption{Per-animal performance in the alternation cohort. Values use the same held-out predictions as \cref{tab:primary_cohort_uncertainty}. Identity is percentage accuracy; the remaining targets are MSE.}
    \label{tab:performance_per_rat}
    \begin{adjustbox}{max width=\linewidth}
      \begin{tabular}{@{}lccccc@{}}
        \toprule
        \textbf{Rat} & \textbf{Identity} ($\uparrow$) & \textbf{Day} ($\downarrow$) & \textbf{Proficiency} ($\downarrow$) & \textbf{Success Rate} ($\downarrow$) & \textbf{Reward} ($\downarrow$) \\
        \midrule
        \textit{Rat A} & 72.70 & 0.72 & .0119 & 0.22 & 0.53 \\
        \textit{Rat B} & 95.92 & 0.91 & .0112 & 0.12 & 0.56 \\
        \textit{Rat C} & 76.45 & 1.49 & .0155 & 0.18 & 0.38 \\
        \textit{Rat D} & 74.40 & 1.97 & .0313 & 0.16 & 0.44 \\
        \textit{Rat E} & 95.68 & 1.33 & .0087 & 0.04 & 1.15 \\
        \textit{Rat F} & 85.15 & 1.02 & .0145 & 0.11 & 0.55 \\
        \bottomrule
      \end{tabular}
    \end{adjustbox}
  \end{table}

  \subsection{Recording Days, Maze Regions, and Movement Motifs}
  \label{app:region_motif}

  \mypara{Maze regions.}
  We divide the Y-maze into three regions: \textit{junction}, where the three arms intersect; \textit{arm segment}, the straight portion of each arm; and \textit{endpoint}, the distal end of each arm.

  \mypara{Movement motifs.}
  We assign five motifs from the Local Pose and Global Trajectory views: \textit{remaining still}, with little translation or postural change; \textit{scanning}, with head or body reorientation while staying in place; \textit{going straight}, with forward travel along an arm; \textit{turning}, with a change of direction between arms through the junction; and \textit{turning back}, with a reversal near an arm endpoint.

  The subgroup analysis uses nonoverlapping 33-frame histories from the primary test recordings. Of the 3,376 annotated histories, 3,261 also have the 33 subsequent frames required by the feature cache. \Cref{tab:primary_subgroup_composition} reports the sample count for each animal and the spread of the Proficiency target in each group. \Cref{tab:primary_subgroup_uncertainty} gives their prediction errors and animal-bootstrap intervals. All groups use the same fitted probes, with a separate regression probe for each animal.

  \begin{table}[!htbp]
    \centering
    \tablestyle
    \caption{Composition and target spread in the primary subgroup analysis. Counts include only scored windows. Rat labels follow \cref{tab:performance_per_rat}. The final column gives the standard deviation of Proficiency across these windows.}
    \label{tab:primary_subgroup_composition}
    \begin{tabular}{lrrrrrrr}
      \toprule
      \textbf{Group} & \textit{Rat A} & \textit{Rat B} & \textit{Rat C} & \textit{Rat D} & \textit{Rat E} & \textit{Rat F} & \textbf{\shortstack{Proficiency\\SD}} \\
      \midrule
      Arm segment & 63 & 97 & 101 & 122 & 96 & 79 & .174 \\
      Arm endpoint & 320 & 449 & 396 & 376 & 404 & 432 & .155 \\
      Junction & 42 & 40 & 72 & 52 & 77 & 43 & .157 \\
      \midrule
      Day 1 & 82 & 158 & 110 & 114 & 120 & 116 & .105 \\
      Day 2 & 84 & 107 & 114 & 109 & 113 & 108 & .165 \\
      Day 3 & 88 & 97 & 115 & 109 & 116 & 111 & .111 \\
      Day 4 & 87 & 112 & 116 & 107 & 114 & 109 & .080 \\
      Day 5 & 84 & 112 & 114 & 111 & 114 & 110 & .093 \\
      \midrule
      Remaining still & 151 & 189 & 227 & 180 & 219 & 255 & .145 \\
      Scanning & 175 & 239 & 171 & 217 & 131 & 147 & .164 \\
      Going straight & 71 & 83 & 128 & 99 & 90 & 72 & .170 \\
      Turning & 6 & 31 & 6 & 10 & 70 & 28 & .100 \\
      Turning back & 22 & 44 & 37 & 44 & 67 & 52 & .149 \\
      \bottomrule
    \end{tabular}
  \end{table}

  \begin{table}[!htbp]
    \centering
    \tablestyle[3pt]
    \caption{Prediction across maze regions, recording days, and movement motifs. Values are pooled estimates, with animal-bootstrap 95\% confidence intervals directly below. Identity is percentage accuracy; Day and Proficiency are MSE. Day prediction is omitted when day defines the group.}
    \label{tab:primary_subgroup_uncertainty}
    \begin{tabular}{@{}lrccc@{}}
      \toprule
      \textbf{Group} & \textbf{$n$} & \textbf{Identity} & \textbf{Day} & \textbf{Proficiency} \\
      \midrule
      \multicolumn{5}{l}{\textit{Maze region}} \\
      Arm segment & 558 & 81.7 & 1.59 & .0211 \\
      &  & $[72.4,\,90.1]$ & $[1.04,\,2.11]$ & $[.0109,\,.0310]$ \\
      \addlinespace[4pt]
      Arm endpoint & 2377 & 83.0 & 1.20 & .0149 \\
      &  & $[74.1,\,91.0]$ & $[0.92,\,1.50]$ & $[.0113,\,.0203]$ \\
      \addlinespace[4pt]
      Junction & 326 & 88.7 & 1.20 & .0135 \\
      &  & $[81.1,\,95.6]$ & $[0.89,\,1.51]$ & $[.0073,\,.0225]$ \\
      \addlinespace[4pt]
      \midrule
      \multicolumn{5}{l}{\textit{Recording day}} \\
      Day 1 & 700 & 79.3 & -- & .0192 \\
      &  & $[68.5,\,87.6]$ & -- & $[.0107,\,.0290]$ \\
      \addlinespace[4pt]
      Day 2 & 635 & 81.9 & -- & .0183 \\
      &  & $[72.6,\,91.2]$ & -- & $[.0112,\,.0269]$ \\
      \addlinespace[4pt]
      Day 3 & 636 & 78.8 & -- & .0164 \\
      &  & $[66.5,\,90.8]$ & -- & $[.0106,\,.0224]$ \\
      \addlinespace[4pt]
      Day 4 & 645 & 89.5 & -- & .0128 \\
      &  & $[84.3,\,94.4]$ & -- & $[.0063,\,.0200]$ \\
      \addlinespace[4pt]
      Day 5 & 645 & 87.6 & -- & .0123 \\
      &  & $[79.5,\,95.4]$ & -- & $[.0062,\,.0218]$ \\
      \addlinespace[4pt]
      \midrule
      \multicolumn{5}{l}{\textit{Movement motif}} \\
      Remaining still & 1221 & 87.1 & 1.11 & .0115 \\
      &  & $[78.8,\,93.6]$ & $[0.85,\,1.36]$ & $[.0083,\,.0164]$ \\
      \addlinespace[4pt]
      Scanning & 1080 & 76.6 & 1.34 & .0195 \\
      &  & $[67.1,\,86.0]$ & $[0.99,\,1.78]$ & $[.0151,\,.0261]$ \\
      \addlinespace[4pt]
      Going straight & 543 & 84.0 & 1.37 & .0184 \\
      &  & $[76.1,\,92.8]$ & $[0.87,\,1.82]$ & $[.0102,\,.0271]$ \\
      \addlinespace[4pt]
      Turning & 151 & 98.0 & 1.07 & .0083 \\
      &  & $[93.5,\,99.5]$ & $[0.92,\,1.44]$ & $[.0055,\,.0145]$ \\
      \addlinespace[4pt]
      Turning back & 266 & 84.2 & 1.56 & .0202 \\
      &  & $[72.0,\,94.0]$ & $[1.22,\,1.97]$ & $[.0111,\,.0305]$ \\
      \addlinespace[4pt]
      \bottomrule
    \end{tabular}
  \end{table}

  \FloatBarrier
  \subsubsection{Probabilistic Rewards}
  \Cref{tab:probabilistic_subgroups} extends the region, day, and motif analysis to the probabilistic-reward cohort. Identity is decoded across all three maze regions and all five recording days. Turning gives the highest identity accuracy and the lowest Day, Proficiency, and Success Rate errors among the movement motifs; Reward error is lowest during straight movement.

  \begin{table}[!htbp]
    \centering
    \tablestyle
    \caption{Prediction across regions, days, and motifs with probabilistic rewards. Identity is percentage accuracy; the remaining targets are MSE. Region and motif definitions follow the alternation analysis in \cref{app:region_motif}. Day prediction is omitted when day defines the group.}
    \label{tab:probabilistic_subgroups}
    \begin{adjustbox}{max width=\linewidth}
      \begin{tabular}{lrrrrr}
        \toprule
        \textbf{Group} & \textbf{Identity} ($\uparrow$) & \textbf{Day} ($\downarrow$) & \textbf{Proficiency} ($\downarrow$) & \textbf{Success Rate} ($\downarrow$) & \textbf{Reward} ($\downarrow$) \\
        \midrule
        \rowcolor[HTML]{F2F5F7}
        \multicolumn{6}{l}{\textit{Maze region}} \\
        Junction & 85.23 & 1.291 & .0070 & 0.05 & 0.608 \\
        Arm segment & 86.00 & 1.444 & .0084 & 0.06 & 0.638 \\
        Arm endpoint & 85.77 & 1.325 & .0068 & 0.05 & 0.697 \\
        \midrule
        \rowcolor[HTML]{F2F5F7}
        \multicolumn{6}{l}{\textit{Recording day}} \\
        Day 1 & 79.24 & -- & .0077 & 0.07 & 0.541 \\
        Day 2 & 82.45 & -- & .0059 & 0.06 & 0.567 \\
        Day 3 & 89.86 & -- & .0063 & 0.04 & 0.641 \\
        Day 4 & 90.17 & -- & .0058 & 0.04 & 0.881 \\
        Day 5 & 87.06 & -- & .0100 & 0.04 & 0.739 \\
        \midrule
        \rowcolor[HTML]{F2F5F7}
        \multicolumn{6}{l}{\textit{Movement motif}} \\
        Remaining still & 87.72 & 1.157 & .0058 & 0.05 & 0.686 \\
        Scanning & 82.67 & 1.752 & .0087 & 0.06 & 0.695 \\
        Going straight & 83.97 & 1.133 & .0074 & 0.04 & 0.590 \\
        Turning & 96.63 & 0.760 & .0033 & 0.01 & 0.716 \\
        Turning back & 89.66 & 1.067 & .0065 & 0.04 & 0.687 \\
        \bottomrule
      \end{tabular}
    \end{adjustbox}
  \end{table}

  \FloatBarrier
  \subsection{Movement Descriptors and Maze Geometry}

  \subsubsection{Movement Descriptors}
  We compare the supplementary three-second representation with mean body--neck and neck--nose lengths, and with 14 kinematic features describing displacement, speed, acceleration, scanning, reversals, backtracking, lateral deviation, and head--body decoupling. Linear regularization is selected on validation data. The regression constant is each animal's training-target mean. TRAIL improves identity recognition over both bone lengths and kinematics, and predicts day and reward frequency more accurately than the kinematic features and training-mean predictor (\cref{tab:additional_simple_controls}).

  \begin{table}[!htbp]
    \centering
    \tablestyle
    \caption{Movement-descriptor comparison. Identity is accuracy; the remaining readouts are MSE. Results use 105{,}882 held-out-time windows. The final rows compare TRAIL with the constant predictor: $R^2$ uses aggregate MSE, and reductions are averaged across animals. Intervals below the reductions are paired animal-bootstrap 95\% confidence intervals.}
    \label{tab:additional_simple_controls}
    \begin{adjustbox}{max width=\linewidth}
      \begin{tabular}{lrrrr}
        \toprule
        \textbf{Representation} & \textbf{Identity} ($\uparrow$) & \textbf{Day} ($\downarrow$) & \textbf{Proficiency} ($\downarrow$) & \textbf{Reward} ($\downarrow$) \\
        \midrule
        Constant / chance & 16.67\% & 2.040 & .02163 & 0.805 \\
        Mean bone lengths & 29.32\% & -- & -- & -- \\
        14 kinematic features & 28.81\% & 1.895 & .02031 & 0.753 \\
        \rowcolor[HTML]{E3ECF5}
        \ours & 84.07\% & 1.252 & .01552 & 0.593 \\
        \midrule
        $R^2$ & -- & 0.386 & .282 & 0.264 \\
        MSE reduction & -- & 0.800 & .00622 & 0.200 \\
         & -- & $[0.429,\,1.115]$ & $[.00268,\,.01014]$ & $[0.0509,\,0.371]$ \\
        \bottomrule
      \end{tabular}
    \end{adjustbox}
  \end{table}

  \subsubsection{Identity from Average Segment Lengths}
  To examine how much identity information is carried by apparent body size, we represent each 33-frame trajectory by two numbers: the mean body--neck length and the mean neck--nose length across its frames. Five classifiers predict identity among the six rats in the alternation cohort from these two features alone. We evaluate the default split and a held-out-day split that trains on days 1--4 and tests on day 5 (\cref{tab:segment_length_identity}). Classification remains above the majority-class baseline in both settings, with similar accuracy on the held-out day. Thus, average segment lengths carry identity information that persists across recording days.

  For kNN, we use uniform neighbor weights with $k=51$ on the default split and $k=151$ on the held-out-day split. The remaining classifier settings are the same for both splits: SVM-RBF uses $C=10$ and \texttt{gamma=scale}; the decision tree and random forest use minimum leaf sizes of 100 and 50, respectively; and TabPFN v2.6 uses its default settings.

  \begin{table}[!htbp]
    \centering
    \tablestyle
    \caption{Identity prediction from two mean segment lengths. Accuracy, balanced accuracy, and macro-F1 are percentages. The default test split contains 3,376 trajectories. Majority-class accuracy is 18.0\% for the default split and 17.7\% for the held-out day.}
    \label{tab:segment_length_identity}
    \begin{adjustbox}{max width=\linewidth}
      \begin{tabular}{lrrr}
        \toprule
        \textbf{Classifier} & \textbf{Accuracy} & \textbf{Balanced acc.} & \textbf{Macro-F1} \\
        \midrule
        \rowcolor[HTML]{F2F5F7}
        \multicolumn{4}{l}{\textit{Default split}} \\
        kNN & 41.4 & 40.1 & 38.4 \\
        SVM-RBF & 41.3 & 39.8 & 35.9 \\
        Decision tree & 40.3 & 38.9 & 37.2 \\
        Random forest & 41.9 & 40.4 & 38.1 \\
        TabPFN v2.6 & 42.3 & 40.8 & 38.1 \\
        \midrule
        \rowcolor[HTML]{F2F5F7}
        \multicolumn{4}{l}{\textit{Held-out day: train on days 1--4, test on day 5}} \\
        kNN & 41.8 & 40.4 & 38.4 \\
        SVM-RBF & 42.6 & 41.0 & 36.7 \\
        Decision tree & 41.0 & 39.8 & 38.6 \\
        Random forest & 41.7 & 40.3 & 38.1 \\
        TabPFN v2.6 & 42.4 & 40.9 & 37.9 \\
        \bottomrule
      \end{tabular}
    \end{adjustbox}
  \end{table}

  \Cref{tab:segment_length_recall} shows the corresponding recall for each rat on the default split. The classifiers show a similar pattern across animals, with some identities more readily distinguished by these two measurements than others.

  \begin{table}[!htbp]
    \centering
    \tablestyle
    \caption{Per-animal recall from mean segment lengths. Values are percentages on the default split in \cref{tab:segment_length_identity}. Rat labels follow \cref{tab:performance_per_rat}.}
    \label{tab:segment_length_recall}
    \begin{tabular}{lrrrrrr}
      \toprule
      \textbf{Classifier} & \textit{Rat A} & \textit{Rat B} & \textit{Rat C} & \textit{Rat D} & \textit{Rat E} & \textit{Rat F} \\
      \midrule
      kNN & 17 & 50 & 21 & 40 & 81 & 32 \\
      SVM-RBF & 6 & 43 & 11 & 44 & 85 & 49 \\
      Decision tree & 12 & 49 & 24 & 37 & 78 & 34 \\
      Random forest & 11 & 49 & 21 & 42 & 84 & 35 \\
      TabPFN v2.6 & 11 & 49 & 16 & 44 & 84 & 42 \\
      \bottomrule
    \end{tabular}
  \end{table}

  \subsubsection{Maze Geometry}
  Adding maze geometry to the body-centered trajectories improves identity recognition and lowers day and reward-frequency error (\cref{tab:additional_maze_context}).

  \begin{table}[!htbp]
    \centering
    \tablestyle
    \caption{Effect of maze context. Identity is accuracy; the remaining readouts are MSE. Both runs use seed 42 and the same 105{,}882 held-out time windows.}
    \label{tab:additional_maze_context}
    \begin{adjustbox}{max width=\linewidth}
      \begin{tabular}{lrrrr}
        \toprule
        \textbf{Maze context} & \textbf{Identity} ($\uparrow$) & \textbf{Day} ($\downarrow$) & \textbf{Proficiency} ($\downarrow$)  & \textbf{Reward} ($\downarrow$) \\
        \midrule
        Removed & 70.88\% & 1.599 & .01994 & 0.648 \\
        \rowcolor[HTML]{E3ECF5}
        Included & 84.07\% & 1.252 & .01552 & 0.593 \\
        \bottomrule
      \end{tabular}
    \end{adjustbox}
  \end{table}

  \FloatBarrier
  \needroom{52\baselineskip}
  \section{Representation Visualizations}
  \Cref{fig:baseline_tsne_all} compares three-second representations from the primary evaluation. Each left panel is a joint t-SNE projection colored by rat; the six right panels are independently fitted per-rat projections colored by recording day. Each projection has its own coordinate system.
  \begin{figure}[H]
    \centering
    \setlength{\parskip}{0pt}
    \captionsetup[subfigure]{skip=3pt}
    \begin{subfigure}[t]{0.48\linewidth}
      \centering
      \makebox[\linewidth][c]{\includegraphics[width=197.29169pt,height=105.6124pt]{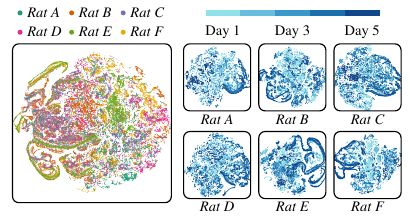}}
      \caption{Direct Trajectory}
      \label{fig:tsne_direct_trajectory}
    \end{subfigure}  \hfill
    \begin{subfigure}[t]{0.48\linewidth}
      \centering
      \makebox[\linewidth][c]{\includegraphics[width=197.29169pt,height=105.6124pt]{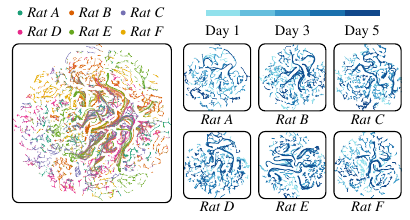}}
      \caption{LG-SSM}
      \label{fig:tsne_lg_ssm}
    \end{subfigure}  \par\smallskip
    \begin{subfigure}[t]{0.48\linewidth}
      \centering
      \makebox[\linewidth][c]{\includegraphics[width=197.29169pt,height=105.6124pt]{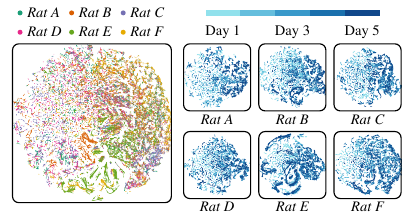}}
      \caption{AR-HMM}
      \label{fig:tsne_ar_hmm}
    \end{subfigure}  \hfill
    \begin{subfigure}[t]{0.48\linewidth}
      \centering
      \makebox[\linewidth][c]{\includegraphics[width=197.29169pt,height=105.6124pt]{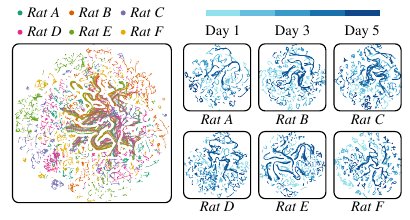}}
      \caption{ProMPs}
      \label{fig:tsne_promps}
    \end{subfigure}  \par\smallskip
    \begin{subfigure}[t]{0.48\linewidth}
      \centering
      \makebox[\linewidth][c]{\includegraphics[width=197.29169pt,height=105.6124pt]{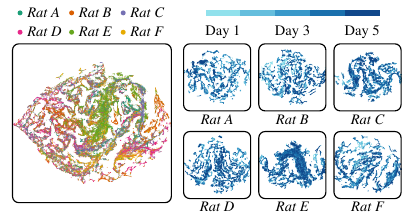}}
      \caption{Keypoint-MoSeq}
      \label{fig:tsne_keypoint_moseq}
    \end{subfigure}  \hfill
    \begin{subfigure}[t]{0.48\linewidth}
      \centering
      \makebox[\linewidth][c]{\includegraphics[width=197.29169pt,height=105.6124pt]{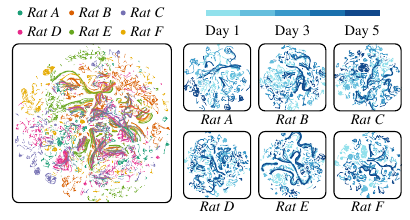}}
      \caption{AutoEncoder}
      \label{fig:tsne_autoencoder}
    \end{subfigure}  \par\smallskip
    \begin{subfigure}[t]{0.48\linewidth}
      \centering
      \makebox[\linewidth][c]{\includegraphics[width=197.29169pt,height=105.6124pt]{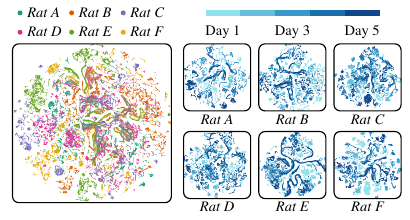}}
      \caption{VAME}
      \label{fig:tsne_vame}
    \end{subfigure}  \hfill
    \begin{subfigure}[t]{0.48\linewidth}
      \centering
      \makebox[\linewidth][c]{\includegraphics[width=197.29169pt,height=105.6124pt]{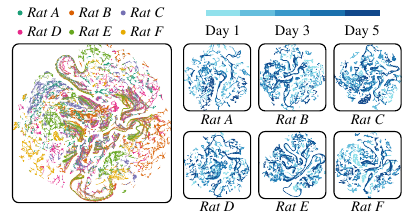}}
      \caption{T-BYOL}
      \label{fig:tsne_t_byol}
    \end{subfigure}  \par
    \caption{t-SNE visualizations of 3-second representations for each baseline. \textit{Left:} A joint projection colored by rat. \textit{Right:} Six per-rat projections colored by day. The figure continues on the next page.}
    \label{fig:baseline_tsne_all}
  \end{figure}
  \begin{figure}[H]
    \ContinuedFloat
    \centering
    \setlength{\parskip}{0pt}
    \captionsetup[subfigure]{skip=3pt}
    \begin{subfigure}[t]{0.48\linewidth}
      \centering
      \makebox[\linewidth][c]{\includegraphics[width=197.29169pt,height=105.6124pt]{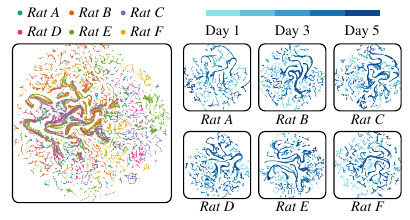}}
      \caption{TCN}
      \label{fig:tsne_tcn}
    \end{subfigure}  \hfill
    \begin{subfigure}[t]{0.48\linewidth}
      \centering
      \makebox[\linewidth][c]{\includegraphics[width=197.29169pt,height=105.6124pt]{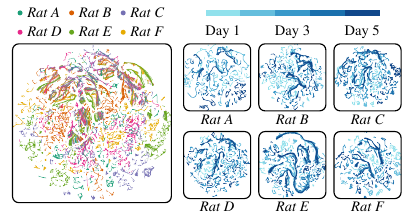}}
      \caption{BAMS}
      \label{fig:tsne_bams}
    \end{subfigure}  \par\smallskip
    \begin{subfigure}[t]{0.48\linewidth}
      \centering
      \makebox[\linewidth][c]{\includegraphics[width=197.29169pt,height=105.6124pt]{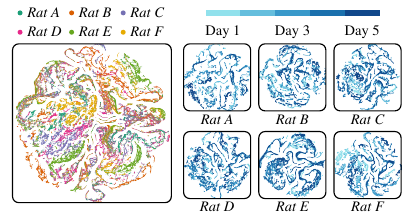}}
      \caption{hBehaveMAE}
      \label{fig:tsne_hbehavemae}
    \end{subfigure}  \hfill
    \begin{subfigure}[t]{0.48\linewidth}
      \centering
      \makebox[\linewidth][c]{\includegraphics[width=197.29169pt,height=105.6124pt]{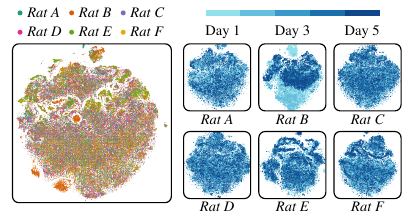}}
      \caption{TimesFM-3 (Zero-shot)}
      \label{fig:tsne_timesfm_zero}
    \end{subfigure}  \par\smallskip
    \begin{subfigure}[t]{0.48\linewidth}
      \centering
      \makebox[\linewidth][c]{\includegraphics[width=197.29169pt,height=105.6124pt]{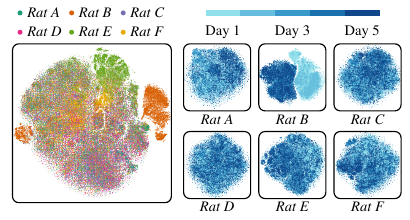}}
      \caption{TimesFM-3 (LoRA)}
      \label{fig:tsne_timesfm_lora}
    \end{subfigure}  \hfill
    \begin{subfigure}[t]{0.48\linewidth}
      \centering
      \makebox[\linewidth][c]{\includegraphics[width=197.29169pt,height=105.6124pt]{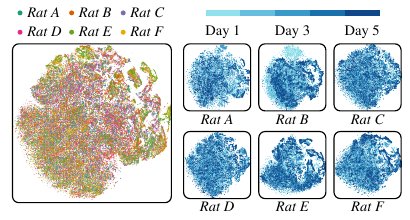}}
      \caption{Chronos-2 (Zero-shot)}
      \label{fig:tsne_chronos_zero}
    \end{subfigure}  \par\smallskip
    \begin{subfigure}[t]{0.48\linewidth}
      \centering
      \makebox[\linewidth][c]{\includegraphics[width=197.29169pt,height=105.6124pt]{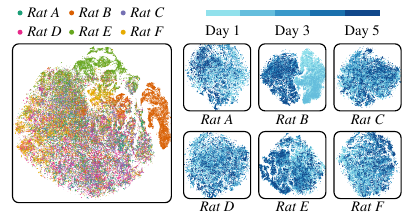}}
      \caption{Chronos-2 (LoRA)}
      \label{fig:tsne_chronos_lora}
    \end{subfigure}  \hfill
    \begin{subfigure}[t]{0.48\linewidth}
      \centering
      \makebox[\linewidth][c]{\includegraphics{fig/tsne-trail.pdf}}
      \caption{\ours}
    \end{subfigure}
    \caption{t-SNE visualizations of 3-second representations (continued). The final panel repeats \ours from \cref{fig:tsne_rat} for direct comparison. Colors follow the same rat and day conventions as above.}
  \end{figure}
  \FloatBarrier
  \section{Movement Transformations}
  \label{app:interpretation}

  \subsection{Estimating Directions of Progression}
  \label[appendix]{app:interpretation_directions}

  \mypara{Centers and axes.}
  Let $\mathcal{S}_{a,d}$ denote the recording sessions for animal $a$ on day $d$, and let $h_{s,i}$ be the representation of window $i$ in session $s$. We first average the $N_s$ representations within each session, then give session means equal weight:
  \begin{equation}
    \mu_s=\frac{1}{N_s}\sum_{i=1}^{N_s}h_{s,i}, \qquad c_{a,d}=\frac{1}{|\mathcal{S}_{a,d}|} \sum_{s\in\mathcal{S}_{a,d}}\mu_s.
  \end{equation}
  Cross-day PCA uses the five centers $\mathcal{C}^{\mathrm{day}}_a=\{c_{a,d}\}_{d=1}^{5}$ of one animal; cross-individual PCA uses the six centers $\mathcal{C}^{\mathrm{individual}}_d=\{c_{a,d}\}_{a=1}^{6}$ on the same relative recording day. For either set $\mathcal{C}$, we define the unit axis by
  \begin{equation}
    v(\mathcal{C})=\underset{\|w\|_2=1}{\operatorname{arg\,max}} \sum_{c\in\mathcal{C}}\bigl[w^\top(c-\bar c)\bigr]^2, \qquad \bar c=\frac{1}{|\mathcal{C}|}\sum_{c\in\mathcal{C}}c.
  \end{equation}
  This is the first principal component of the mean-centered daily centers, without dimension-wise standardization. Each animal has its own cross-day axis and each day its own cross-individual axis; these remain fixed across trajectory-pair comparisons.

  \mypara{Reference-directed transformation.}
  The observed and reference representations determine the direction and extent of movement along $v$ as in \cref{eq:representation_path}. We evaluate this path at $\tau\in\{0,0.1,\ldots,1\}$. At $\tau=1$, the condition matches the reference's projection onto the axis while retaining the observed representation's orthogonal components. Choosing a later-day reference traces variation toward a later stage of the same animal's experience, while choosing another animal on the same relative day traces individual variation. Decoding these conditions with the fixed diffusion latent moves changes associated with each direction observable.

  \subsection{Complementary Views of Turning Transformations}
  \label{app:interpretation_complementary}
  The complementary views in \cref{fig:interpretation_cross_day_local_pose,fig:interpretation_cross_identity_body_trajectory} show how the same turning transformations appear at the postural and trajectory levels, respectively, extending the observations in \cref{fig:interpretation_progression_turning}.

  \mypara{Cross-day transformation of postural timing.}
  The Local Pose view in \cref{fig:interpretation_cross_day_local_pose} reveals a temporal shift within a largely preserved pattern of body-segment proportions. The transformed body-ratio curve closely follows the observed and reconstructed curves, while the skeleton sequence advances through similar configurations earlier. The head's rightward inclination and subsequent realignment, for example, occur earlier in the transformed row than in the reconstructed row. The prominent change is therefore in the timing of postural transitions, with little change in body ratio. This temporal shift complements the changes in speed and turning rhythm in \cref{fig:interpretation_progression_turning}, connecting the trajectory-level observation to the progression of local posture through the turn.

  \setlength{\intextsep}{8pt}
  \resultfigure
    {\includegraphics[width=238.48663pt,height=208.62096pt]{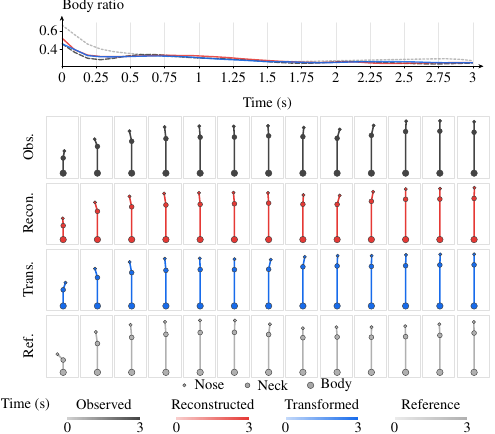}}
    {Cross-day transformation of turning in local pose. Skeleton sequences and body ratio for the same transformation as the left panel of \cref{fig:interpretation_progression_turning}.}
    {fig:interpretation_cross_day_local_pose}

  \mypara{Cross-individual transformation of movement pace.}
  In \cref{fig:interpretation_cross_identity_body_trajectory}, the reference animal's movement characteristics become apparent in the temporal organization of body travel. Relative to the reconstructed movement, the transformed sequence exhibits a lower, later speed peak and maintains forward movement toward the end of the window, closely following the reference speed curve. The turn thus adopts a more gradual progression and a less pronounced terminal slowdown. Together with the changes in body extension in \cref{fig:interpretation_progression_turning}, this complementary view shows the transformation expressing individual differences through both postural dynamics and movement pace.

  \resultfigure
    {\includegraphics[width=238.48663pt,height=208.62096pt]{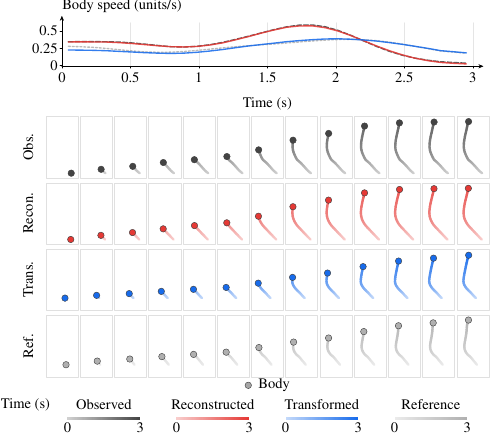}}
    {Cross-individual transformation of turning in body trajectory. Body-keypoint trajectories and body speed for the same transformation as the right panel of \cref{fig:interpretation_progression_turning}.}
    {fig:interpretation_cross_identity_body_trajectory}

  \FloatBarrier
  \subsection{Movement Transformations in Going Straight}
  \label[appendix]{app:interpretation_straight}
  Going straight provides a complementary setting in which changes in movement execution can be followed along a largely consistent direction of travel. \Cref{fig:interpretation_cross_day_straight,fig:interpretation_cross_identity_straight} present cross-day and cross-individual transformations, respectively, with Body Trajectory and body speed on the left and Local Pose and body ratio on the right.

  Across days, the transformation reorganizes when movement is concentrated within the sequence. In \cref{fig:interpretation_cross_day_straight}, the observed and reconstructed movements begin relatively slowly and accelerate again later in the window. The transformed movement instead starts faster and, after an early peak, progressively slows toward the end, taking on the later-day reference's pattern of movement. The Local Pose sequences retain a broadly similar pattern of body extension, accompanied by closely aligned reconstructed and transformed body-ratio curves. As in the turning example, the transformation expresses day-related variation primarily through changes in pace and timing while preserving the broad spatial structure of the movement.

  \resultfigure
    {\includegraphics[width=349.78455pt,height=161.77739pt]{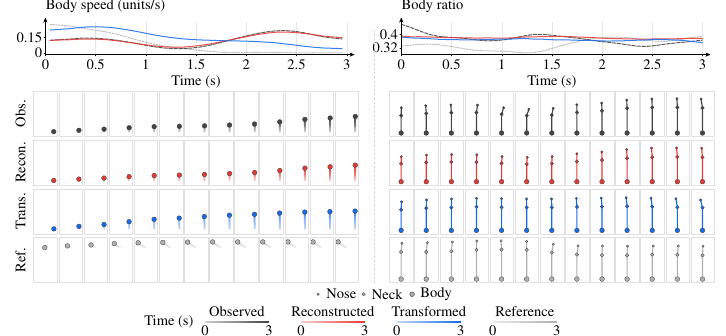}}
    {Cross-day transformation of going straight. \textit{Left:} Body Trajectory with body speed. \textit{Right:} Local Pose with body ratio.}
    {fig:interpretation_cross_day_straight}

  Across individuals, \cref{fig:interpretation_cross_identity_straight} shows a shift toward the reference in both movement pace and the relative extension of the tracked body segments. The transformed body-speed curve approaches the reference's earlier peak and subsequent decline. In Local Pose, the transformed sequence maintains a higher nose--neck to neck--body ratio through much of the middle of the movement, bringing it closer to the reference than the reconstructed sequence. These changes extend the observation from turning: individual characteristics become visible through both how movement unfolds over time and how the body extends during its execution.

  \resultfigure
    {\includegraphics[width=349.78455pt,height=161.77739pt]{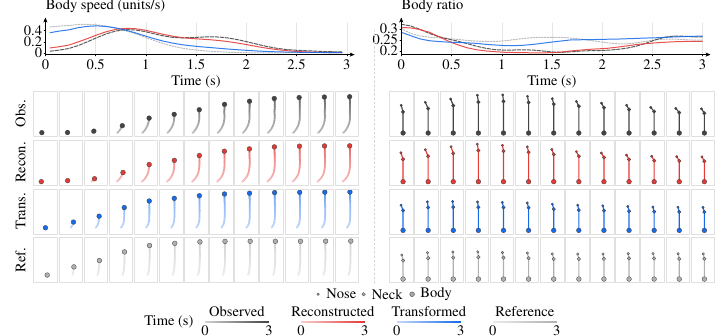}}
    {Cross-individual transformation of going straight. \textit{Left:} Body Trajectory with body speed. \textit{Right:} Local Pose with body ratio.}
    {fig:interpretation_cross_identity_straight}

  \FloatBarrier
  \subsection{Diffusion Model and Training Details}
  \label[appendix]{app:interpretation_implementation}

  \mypara{Data and backbone.}
  We train the diffusion model using the existing training and validation splits from the six alternation-task animals (\textit{Rat A}--\textit{Rat F}). Each 3-second trajectory contains 33 frames of two-dimensional nose, neck, and body coordinates in the preprocessed absolute coordinate system, paired with a 16-dimensional representation $h$ extracted by the frozen movement encoder. The backbone follows DiT-B~\citep{peebles2023scalable}, with 12 transformer blocks, hidden dimension 768, 12 attention heads, and an MLP expansion ratio of 4. Each keypoint at each frame forms a token. The representation $h$ and diffusion timestep are incorporated through adaLN-Zero, and the model predicts the added noise and reverse-process variance.

  \mypara{Denoising and training objective.}
  For each training trajectory $x_0$, we uniformly sample a diffusion timestep $t$ and construct a noisy trajectory:
  \begin{equation}
    x_t=\sqrt{\bar\alpha_t}\,x_0 +\sqrt{1-\bar\alpha_t}\,\epsilon, \qquad \epsilon\sim\mathcal{N}(0,I).
  \end{equation}
  We use 1000 diffusion steps with a linear variance schedule from $10^{-4}$ to $0.02$, where $\bar\alpha_t$ is the cumulative signal-retention factor. Conditioned on $h$, the model learns to recover the added noise from $x_t$ while estimating the reverse-process variance:
  \begin{equation}
    \mathcal{L}_{\mathrm{diff}} = \mathbb{E}_{x_0,h,t,\epsilon}\!\left[ \operatorname{MSE}\!\bigl(\epsilon,\epsilon_\psi(x_t,t,h)\bigr) + \mathcal{L}_{\mathrm{var}} \right],
  \end{equation}
  \begin{equation}
    \mathcal{L}_{\mathrm{var}} = \frac{1}{\ln 2}\, \operatorname{mean}_{j}\, D_{\mathrm{KL}}\!\left( q(x_{t-1,j}\mid x_t,x_0) \,\Vert\, p_\psi(x_{t-1,j}\mid x_t,h) \right), \qquad t>0.
  \end{equation}
  The MSE term measures noise-prediction error, while the KL term trains the reverse-process variance by comparing the forward-process posterior $q$ with the learned reverse transition $p_\psi$. The factor $1/\ln 2$ expresses the divergence in bits. Both terms are averaged over coordinates $j$ in frames with no missing keypoints. For variance learning, gradients do not propagate through the predicted noise, and $\mathcal{L}_{\mathrm{var}}$ is set to zero at $t=0$.

  \mypara{Optimization.}
  With the movement encoder frozen, we train the diffusion model from random initialization for 100 epochs using AdamW, a learning rate of $10^{-4}$, batch size 256 per device, and zero weight decay. We maintain an exponential moving average (EMA) of model parameters with decay 0.9999 and select the EMA checkpoint with the lowest validation denoising loss.

  \mypara{DDIM inversion and generation.}
  We use deterministic DDIM~\citep{song2020denoising} with 100 steps and $\eta=0$. DDIM inversion maps an observed trajectory $x_o$ to a diffusion latent $u_o$ by approximately reversing the denoising process. With the original representation $h_o$ fixed, each step estimates the clean trajectory and advances to a higher noise level:
  \begin{equation}
    \begin{aligned}
      \widehat{x}_0 &= \frac{x_t-\sqrt{1-\bar\alpha_t}\, \epsilon_\psi(x_t,t,h_o)} {\sqrt{\bar\alpha_t}},\\
      x_{t'} &= \sqrt{\bar\alpha_{t'}}\,\widehat{x}_0 + \sqrt{1-\bar\alpha_{t'}}\, \epsilon_\psi(x_t,t,h_o), \qquad t'>t.
    \end{aligned}
  \end{equation}
  Here, $t'$ is the next timestep on the inversion grid. Starting from the observed trajectory, these updates progressively increase the noise level, defining the approximate inversion mapping $u_o=\mathcal{I}_\psi(x_o,h_o)$. Denoising this latent under the original condition gives the reconstruction. We then hold $u_o$ and the model parameters fixed while varying the condition along the path in \cref{eq:representation_path}.


\clearpage
  \begin{thebibliography}{43}
    \providecommand{\doi}{doi: \begingroup \urlstyle{rm}\Url}

    \bibitem[Ansari et~al.(2025)Ansari, Shchur, Küken, Auer, Han, Mercado,
    Rangapuram, Shen, Stella, Zhang, Goswami, Kapoor, Maddix, Guerron, Hu, Yin,
    Erickson, Desai, Wang, Rangwala, Karypis, Wang, and
    Bohlke-Schneider]{ansari2025chronos2}
    Abdul~Fatir Ansari, Oleksandr Shchur, Jaris Küken, Andreas Auer, Boran Han,
    Pedro Mercado, Syama~Sundar Rangapuram, Huibin Shen, Lorenzo Stella, Xiyuan
    Zhang, Mononito Goswami, Shubham Kapoor, Danielle~C. Maddix, Pablo Guerron,
    Tony Hu, Junming Yin, Nick Erickson, Prateek~Mutalik Desai, Hao Wang, Huzefa
    Rangwala, George Karypis, Yuyang Wang, and Michael Bohlke-Schneider.
    \newblock Chronos-2: From univariate to universal forecasting.
    \newblock \emph{arXiv preprint arXiv:2510.15821}, 2025.
    \newblock URL \url{https://arxiv.org/abs/2510.15821}.

    \bibitem[Azabou et~al.(2023)Azabou, Mendelson, Ahad, Sorokin, Thakoor, Urzay,
    and Dyer]{NEURIPS2023_5aad86aa}
    Mehdi Azabou, Michael Mendelson, Nauman Ahad, Maks Sorokin, Shantanu Thakoor,
    Carolina Urzay, and Eva Dyer.
    \newblock Relax, it doesn't matter how you get there: A new self-supervised
    approach for multi-timescale behavior analysis.
    \newblock In A.~Oh, T.~Naumann, A.~Globerson, K.~Saenko, M.~Hardt, and
    S.~Levine (eds.), \emph{Advances in Neural Information Processing Systems},
    volume~36, pp.\  28491--28509. Curran Associates, Inc., 2023.
    \newblock \doi{10.52202/075280-1238}.
    \newblock URL
    \url{https://proceedings.neurips.cc/paper_files/paper/2023/file/5aad86aa2a3c00b70c71e19bc4780319-Paper-Conference.pdf}.

    \bibitem[Bai et~al.(2018)Bai, Kolter, and Koltun]{BaiTCN2018}
    Shaojie Bai, J.~Zico Kolter, and Vladlen Koltun.
    \newblock An empirical evaluation of generic convolutional and recurrent
    networks for sequence modeling.
    \newblock \emph{arXiv:1803.01271}, 2018.

    \bibitem[Bova et~al.(2021)Bova, Ferris, and Leventhal]{bova2021reaching}
    Alexandra Bova, Kenneth Ferris, and Daniel~K. Leventhal.
    \newblock Evolution of gross forelimb and fine digit kinematics during skilled
    reaching acquisition in rats.
    \newblock \emph{eNeuro}, 8\penalty0 (5):\penalty0 ENEURO.0153--21.2021, 2021.
    \newblock \doi{10.1523/ENEURO.0153-21.2021}.

    \bibitem[Chao et~al.(2019)Chao, He, Zhang, and Feng]{chao2019gaitset}
    Hanqing Chao, Yiwei He, Junping Zhang, and Jianfeng Feng.
    \newblock {GaitSet}: Regarding gait as a set for cross-view gait recognition.
    \newblock In \emph{Proceedings of the AAAI Conference on Artificial
      Intelligence}, volume~33, pp.\  8126--8133, 2019.
    \newblock \doi{10.1609/aaai.v33i01.33018126}.

    \bibitem[Das et~al.(2024)Das, Kong, Sen, and Zhou]{pmlr-v235-das24c}
    Abhimanyu Das, Weihao Kong, Rajat Sen, and Yichen Zhou.
    \newblock A decoder-only foundation model for time-series forecasting.
    \newblock In Ruslan Salakhutdinov, Zico Kolter, Katherine Heller, Adrian
    Weller, Nuria Oliver, Jonathan Scarlett, and Felix Berkenkamp (eds.),
    \emph{Proceedings of the 41st International Conference on Machine Learning},
    volume 235 of \emph{Proceedings of Machine Learning Research}, pp.\
    10148--10167. PMLR, 21--27 Jul 2024.
    \newblock URL \url{https://proceedings.mlr.press/v235/das24c.html}.

    \bibitem[Dhawale et~al.(2019)Dhawale, Miyamoto, Smith, and
    {\"O}lveczky]{dhawale2019adaptive}
    Ashesh~K. Dhawale, Yohsuke~R. Miyamoto, Maurice~A. Smith, and Bence~P.
    {\"O}lveczky.
    \newblock Adaptive regulation of motor variability.
    \newblock \emph{Current Biology}, 29\penalty0 (21):\penalty0 3551--3562, 2019.
    \newblock \doi{10.1016/j.cub.2019.08.052}.

    \bibitem[Ding et~al.(2026)Ding, Gao, Qian, and Cui]{ding2026geese}
    Yiran Ding, Yuen Gao, Chunqi Qian, and Zijun Cui.
    \newblock {GEESE}: Genotype-aware end-to-end spatio-temporal embedding for
    behavioral phenotyping, 2026.

    \bibitem[Durbin \& Koopman(2012)Durbin and Koopman]{Durbin_2012}
    James Durbin and Siem~Jan Koopman.
    \newblock \emph{Time Series Analysis by State Space Methods}.
    \newblock Oxford University Press, second edition, May 2012.
    \newblock ISBN 9780199641178.
    \newblock \doi{10.1093/acprof:oso/9780199641178.001.0001}.
    \newblock URL
    \url{http://dx.doi.org/10.1093/acprof:oso/9780199641178.001.0001}.

    \bibitem[Forkosh et~al.(2019)Forkosh, Karamihalev, Roeh, Alon, Anpilov, Touma,
    Nussbaumer, Flachskamm, Kaplick, Shemesh, and Chen]{forkosh2019identity}
    Oren Forkosh, Stoyo Karamihalev, Simone Roeh, Uri Alon, Sergey Anpilov, Chadi
    Touma, Markus Nussbaumer, Cornelia Flachskamm, Paul~M. Kaplick, Yair Shemesh,
    and Alon Chen.
    \newblock Identity domains capture individual differences from across the
    behavioral repertoire.
    \newblock \emph{Nature Neuroscience}, 22\penalty0 (12):\penalty0 2023--2028,
    2019.
    \newblock \doi{10.1038/s41593-019-0516-y}.

    \bibitem[Freund et~al.(2013)Freund, Brandmaier, Lewejohann, Kirste, Kritzler,
    Kr{\"u}ger, Sachser, Lindenberger, and Kempermann]{freund2013individuality}
    Julia Freund, Andreas~M. Brandmaier, Lars Lewejohann, Imke Kirste, Mareike
    Kritzler, Antonio Kr{\"u}ger, Norbert Sachser, Ulman Lindenberger, and Gerd
    Kempermann.
    \newblock Emergence of individuality in genetically identical mice.
    \newblock \emph{Science}, 340\penalty0 (6133):\penalty0 756--759, 2013.
    \newblock \doi{10.1126/science.1235294}.

    \bibitem[Godesberg et~al.(2024)Godesberg, Bockem{\"u}hl, and
    B{\"u}schges]{godesberg2024walking}
    Vincent Godesberg, Till Bockem{\"u}hl, and Ansgar B{\"u}schges.
    \newblock Natural variability and individuality of walking behavior in
    {Drosophila}.
    \newblock \emph{Journal of Experimental Biology}, 227\penalty0 (22):\penalty0
    jeb247878, 2024.
    \newblock \doi{10.1242/jeb.247878}.

    \bibitem[{Google Research}(2026)]{google2026timesfm3}
    {Google Research}.
    \newblock {TimesFM} 3.0 ({PyTorch}).
    \newblock Official model card, 2026.
    \newblock URL \url{https://huggingface.co/google/timesfm-3.0-pytorch}.

    \bibitem[Grill et~al.(2020)Grill, Strub, Altch\'{e}, Tallec, Richemond,
    Buchatskaya, Doersch, Avila~Pires, Guo, Gheshlaghi~Azar, Piot, Kavukcuoglu,
    Munos, and Valko]{NEURIPS2020_f3ada80d}
    Jean-Bastien Grill, Florian Strub, Florent Altch\'{e}, Corentin Tallec, Pierre
    Richemond, Elena Buchatskaya, Carl Doersch, Bernardo Avila~Pires, Zhaohan
    Guo, Mohammad Gheshlaghi~Azar, Bilal Piot, Koray Kavukcuoglu, Remi Munos, and
    Michal Valko.
    \newblock Bootstrap your own latent - a new approach to self-supervised
    learning.
    \newblock In H.~Larochelle, M.~Ranzato, R.~Hadsell, M.F. Balcan, and H.~Lin
    (eds.), \emph{Advances in Neural Information Processing Systems}, volume~33,
    pp.\  21271--21284. Curran Associates, Inc., 2020.
    \newblock URL
    \url{https://proceedings.neurips.cc/paper_files/paper/2020/file/f3ada80d5c4ee70142b17b8192b2958e-Paper.pdf}.

    \bibitem[Hinton \& Salakhutdinov(2006)Hinton and Salakhutdinov]{Hinton_2006}
    G.~E. Hinton and R.~R. Salakhutdinov.
    \newblock Reducing the dimensionality of data with neural networks.
    \newblock \emph{Science}, 313\penalty0 (5786):\penalty0 504--507, July 2006.
    \newblock ISSN 1095-9203.
    \newblock \doi{10.1126/science.1127647}.
    \newblock URL \url{http://dx.doi.org/10.1126/science.1127647}.

    \bibitem[Hu et~al.(2022)Hu, Shen, Wallis, Allen-Zhu, Li, Wang, Wang, and
    Chen]{hu2022lora}
    Edward~J Hu, Yelong Shen, Phillip Wallis, Zeyuan Allen-Zhu, Yuanzhi Li, Shean
    Wang, Lu~Wang, and Weizhu Chen.
    \newblock Lo{RA}: Low-rank adaptation of large language models.
    \newblock In \emph{International Conference on Learning Representations}, 2022.
    \newblock URL \url{https://openreview.net/forum?id=nZeVKeeFYf9}.

    \bibitem[Jadhav \& Frank(2020)Jadhav and Frank]{crcns_hc28}
    Shantanu~P. Jadhav and Loren~M. Frank.
    \newblock Simultaneous extracellular recordings from hippocampal area ca1 and
    medial prefrontal cortex from rats performing a w-track alternation task.,
    2020.
    \newblock URL \url{http://crcns.org/data-sets/hc/hc-28}.

    \bibitem[Kawai et~al.(2015)Kawai, Markman, Poddar, Ko, Fantana, Dhawale,
    Kampff, and {\"O}lveczky]{kawai2015motor}
    Risa Kawai, Timothy Markman, Rajesh Poddar, Raymond Ko, Antoniu~L. Fantana,
    Ashesh~K. Dhawale, Adam~R. Kampff, and Bence~P. {\"O}lveczky.
    \newblock Motor cortex is required for learning but not for executing a motor
    skill.
    \newblock \emph{Neuron}, 86\penalty0 (3):\penalty0 800--812, 2015.
    \newblock \doi{10.1016/j.neuron.2015.03.024}.

    \bibitem[Levy et~al.(2023)Levy, Hunter, Lin, Robinson, Gillis, Conlin, Anyoha,
    Shansky, and Datta]{levy2023individual}
    Dana~Rubi Levy, Nigel Hunter, Sherry Lin, Emma~Marie Robinson, Winthrop Gillis,
    Eli~Benjamin Conlin, Rockwell Anyoha, Rebecca~M. Shansky, and Sandeep~Robert
    Datta.
    \newblock Mouse spontaneous behavior reflects individual variation rather than
    estrous state.
    \newblock \emph{Current Biology}, 33\penalty0 (7):\penalty0 1358--1364, 2023.
    \newblock \doi{10.1016/j.cub.2023.02.035}.

    \bibitem[Li et~al.(2023)Li, Li, Savarese, and Hoi]{li2023blip2}
    Junnan Li, Dongxu Li, Silvio Savarese, and Steven Hoi.
    \newblock Blip-2: Bootstrapping language-image pre-training with frozen image
    encoders and large language models, 2023.
    \newblock URL \url{https://arxiv.org/abs/2301.12597}.

    \bibitem[Luxem et~al.(2022)Luxem, Mocellin, Fuhrmann, K{\"u}rsch, Miller,
    Palop, Remy, and Bauer]{Luxem2022}
    Kevin Luxem, Petra Mocellin, Falko Fuhrmann, Johannes K{\"u}rsch, Stephanie~R.
    Miller, Jorge~J. Palop, Stefan Remy, and Pavol Bauer.
    \newblock Identifying behavioral structure from deep variational embeddings of
    animal motion.
    \newblock \emph{Communications Biology}, 5\penalty0 (1):\penalty0 1267, Nov
    2022.
    \newblock ISSN 2399-3642.
    \newblock \doi{10.1038/s42003-022-04080-7}.
    \newblock URL \url{https://doi.org/10.1038/s42003-022-04080-7}.

    \bibitem[Markowitz et~al.(2023)Markowitz, Gillis, Jay, Wood, Harris,
    Cieszkowski, Scott, Brann, Koveal, Kula, et~al.]{markowitz2023reinforcement}
    Jeffrey~E Markowitz, Winthrop~F Gillis, Maya Jay, Jeffrey Wood, Ryley~W Harris,
    Robert Cieszkowski, Rebecca Scott, David Brann, Dorothy Koveal, Tomasz Kula,
    et~al.
    \newblock Spontaneous behavior is structured by reinforcement without explicit
    reward.
    \newblock \emph{Nature}, 614\penalty0 (7946):\penalty0 108--117, 2023.

    \bibitem[Mathis et~al.(2018)Mathis, Mamidanna, Cury, Abe, Murthy, Mathis, and
    Bethge]{mathis2018deeplabcut}
    Alexander Mathis, Pranav Mamidanna, Kevin~M. Cury, Taiga Abe, Venkatesh~N.
    Murthy, Mackenzie~Weygandt Mathis, and Matthias Bethge.
    \newblock {DeepLabCut}: Markerless pose estimation of user-defined body parts
    with deep learning.
    \newblock \emph{Nature Neuroscience}, 21:\penalty0 1281--1289, 2018.
    \newblock \doi{10.1038/s41593-018-0209-y}.

    \bibitem[Mildenhall et~al.(2022)Mildenhall, Srinivasan, Tancik, Barron,
    Ramamoorthi, and Ng]{mildenhall2022nerf}
    Ben Mildenhall, Pratul~P Srinivasan, Matthew Tancik, Jonathan~T Barron, Ravi
    Ramamoorthi, and Ren Ng.
    \newblock Nerf: Representing scenes as neural radiance fields for view
    synthesis.
    \newblock \emph{Communications of the ACM}, 65\penalty0 (1):\penalty0 99--106,
    2022.
    \newblock \doi{10.1145/3503250}.

    \bibitem[Nayakanti et~al.(2022)Nayakanti, Al-Rfou, Zhou, Goel, Refaat, and
    Sapp]{nayakanti2022wayformer}
    Nigamaa Nayakanti, Rami Al-Rfou, Aurick Zhou, Kratarth Goel, Khaled~S Refaat,
    and Benjamin Sapp.
    \newblock Wayformer: Motion forecasting via simple \& efficient attention
    networks.
    \newblock \emph{arXiv preprint arXiv:2207.05844}, 2022.

    \bibitem[Nguyen et~al.(2021)Nguyen, Sharma, Gil-Silva, Gittis, and
    Chase]{nguyen2021kinematic}
    Katrina~P. Nguyen, Abhinav Sharma, Mauricio Gil-Silva, Aryn~H. Gittis, and
    Steven~M. Chase.
    \newblock Distinct kinematic adjustments over multiple timescales accompany
    locomotor skill development in mice.
    \newblock \emph{Neuroscience}, 466:\penalty0 260--272, 2021.
    \newblock \doi{10.1016/j.neuroscience.2021.05.002}.

    \bibitem[Oord et~al.(2018)Oord, Li, and Vinyals]{oord2018representation}
    Aaron van~den Oord, Yazhe Li, and Oriol Vinyals.
    \newblock Representation learning with contrastive predictive coding.
    \newblock \emph{arXiv preprint arXiv:1807.03748}, 2018.

    \bibitem[Paraschos et~al.(2013)Paraschos, Daniel, Peters, and
    Neumann]{NIPS2013_e53a0a29}
    Alexandros Paraschos, Christian Daniel, Jan~R Peters, and Gerhard Neumann.
    \newblock Probabilistic movement primitives.
    \newblock In C.J. Burges, L.~Bottou, M.~Welling, Z.~Ghahramani, and
    K.~Weinberger (eds.), \emph{Advances in Neural Information Processing
      Systems}, volume~26. Curran Associates, Inc., 2013.
    \newblock URL
    \url{https://proceedings.neurips.cc/paper_files/paper/2013/file/e53a0a2978c28872a4505bdb51db06dc-Paper.pdf}.

    \bibitem[Peebles \& Xie(2023)Peebles and Xie]{peebles2023scalable}
    William Peebles and Saining Xie.
    \newblock Scalable diffusion models with transformers.
    \newblock In \emph{2023 IEEE/CVF International Conference on Computer Vision
      (ICCV)}, pp.\  4172--4182. IEEE, 2023.

    \bibitem[Rosenberg et~al.(2021)Rosenberg, Zhang, Perona, and
    Meister]{rosenberg2021labyrinth}
    Matthew Rosenberg, Tony Zhang, Pietro Perona, and Markus Meister.
    \newblock Mice in a labyrinth show rapid learning, sudden insight, and
    efficient exploration.
    \newblock \emph{eLife}, 10:\penalty0 e66175, 2021.
    \newblock \doi{10.7554/eLife.66175}.

    \bibitem[Schneider et~al.(2023)Schneider, Lee, and Mathis]{schneider2023cebra}
    Steffen Schneider, Jin~Hwa Lee, and Mackenzie~Weygandt Mathis.
    \newblock Learnable latent embeddings for joint behavioral and neural
    analysis.
    \newblock \emph{Nature}, 617\penalty0 (7960):\penalty0 360--368, 2023.
    \newblock \doi{10.1038/s41586-023-06031-6}.

    \bibitem[Sheppard et~al.(2022)Sheppard, Gardin, Sabnis, Peer, Darrell, Deats,
    Geuther, Lutz, and Kumar]{sheppard2022stride}
    Keith Sheppard, Justin Gardin, Gautam~S. Sabnis, Asaf Peer, Megan Darrell, Sean
    Deats, Brian Geuther, Cathleen~M. Lutz, and Vivek Kumar.
    \newblock Stride-level analysis of mouse open field behavior using
    deep-learning-based pose estimation.
    \newblock \emph{Cell Reports}, 38\penalty0 (2):\penalty0 110231, 2022.
    \newblock \doi{10.1016/j.celrep.2021.110231}.

    \bibitem[Shridhar et~al.(2026)Shridhar, Leveroni, Zhou, Quattrocchi, Kaye,
    Gutierrez, Rodr{\'\i}guez~Le{\'o}n, Kay, and Yu]{ShridharENEURO.0048-26.2026}
    Sameera Shridhar, Zachary~M. Leveroni, Nan Zhou, Ellie Quattrocchi, Audrey
    Kaye, Jaquelin Gutierrez, Amanda Rodr{\'\i}guez~Le{\'o}n, Leslie~M. Kay, and
    Jai~Y. Yu.
    \newblock Goal location beta oscillations and sharp wave ripples are
    complementary markers of learning-related changes in hippocampal-neocortical
    networks.
    \newblock \emph{eNeuro}, 2026.
    \newblock \doi{10.1523/ENEURO.0048-26.2026}.
    \newblock URL \url{https://doi.org/10.1523/ENEURO.0048-26.2026}.

    \bibitem[Smith et~al.(2004)Smith, Frank, Wirth, Yanike, Hu, Kubota, Graybiel,
    Suzuki, and Brown]{smith2004dynamic}
    Anne~C. Smith, Loren~M. Frank, Sylvia Wirth, Marianna Yanike, Dan Hu, Yasuo
    Kubota, Ann~M. Graybiel, Wendy~A. Suzuki, and Emery~N. Brown.
    \newblock Dynamic analysis of learning in behavioral experiments.
    \newblock \emph{Journal of Neuroscience}, 24\penalty0 (2):\penalty0 447--461,
    2004.
    \newblock \doi{10.1523/JNEUROSCI.2908-03.2004}.

    \bibitem[Song et~al.(2020)Song, Meng, and Ermon]{song2020denoising}
    Jiaming Song, Chenlin Meng, and Stefano Ermon.
    \newblock Denoising diffusion implicit models.
    \newblock \emph{arXiv preprint arXiv:2010.02502}, 2020.

    \bibitem[Stoffl et~al.(2025)Stoffl, Bonnetto, d'Ascoli, and
    Mathis]{Stoffl2025hBehaveMAE}
    Lucas Stoffl, Andy Bonnetto, St{\'e}phane d'Ascoli, and Alexander Mathis.
    \newblock Elucidating the hierarchical nature of behavior with masked
    autoencoders.
    \newblock In Ale{\v{s}} Leonardis, Elisa Ricci, Stefan Roth, Olga Russakovsky,
    Torsten Sattler, and G{\"u}l Varol (eds.), \emph{Computer Vision -- ECCV
      2024}, pp.\  106--125, Cham, 2025. Springer Nature Switzerland.
    \newblock ISBN 978-3-031-73039-9.
    \newblock \doi{10.1007/978-3-031-73039-9_7}.

    \bibitem[Sun et~al.(2021)Sun, Karigo, Chakraborty, Mohanty, Wild, Sun, Chen,
    Anderson, Perona, Yue, and Kennedy]{calms21}
    Jennifer~J Sun, Tomomi Karigo, Dipam Chakraborty, Sharada Mohanty, Benjamin
    Wild, Quan Sun, Chen Chen, David Anderson, Pietro Perona, Yisong Yue, and Ann
    Kennedy.
    \newblock The multi-agent behavior dataset: Mouse dyadic social interactions.
    \newblock In J.~Vanschoren and S.~Yeung (eds.), \emph{Proceedings of the Neural
      Information Processing Systems Track on Datasets and Benchmarks}, volume~1,
    2021.
    \newblock URL
    \url{https://datasets-benchmarks-proceedings.neurips.cc/paper_files/paper/2021/file/7f1de29e6da19d22b51c68001e7e0e54-Paper-round1.pdf}.

    \bibitem[Sun et~al.(2023)Sun, Marks, Ulmer, Chakraborty, Geuther, Hayes, Jia,
    Kumar, Oleszko, Partridge, Peelman, Robie, Schretter, Sheppard, Sun,
    Uttarwar, Wagner, Werner, Parker, Perona, Yue, Branson, and
    Kennedy]{pmlr-v202-sun23g}
    Jennifer~J. Sun, Markus Marks, Andrew~Wesley Ulmer, Dipam Chakraborty, Brian
    Geuther, Edward Hayes, Heng Jia, Vivek Kumar, Sebastian Oleszko, Zachary
    Partridge, Milan Peelman, Alice Robie, Catherine~E Schretter, Keith Sheppard,
    Chao Sun, Param Uttarwar, Julian~Morgan Wagner, Erik Werner, Joseph Parker,
    Pietro Perona, Yisong Yue, Kristin Branson, and Ann Kennedy.
    \newblock {MAB}e22: A multi-species multi-task benchmark for learned
    representations of behavior.
    \newblock In Andreas Krause, Emma Brunskill, Kyunghyun Cho, Barbara Engelhardt,
    Sivan Sabato, and Jonathan Scarlett (eds.), \emph{Proceedings of the 40th
      International Conference on Machine Learning}, volume 202 of
    \emph{Proceedings of Machine Learning Research}, pp.\  32936--32990. PMLR,
    23--29 Jul 2023.
    \newblock URL \url{https://proceedings.mlr.press/v202/sun23g.html}.

    \bibitem[Tonekaboni et~al.(2021)Tonekaboni, Eytan, and
    Goldenberg]{tonekaboni2021tnc}
    Sana Tonekaboni, Danny Eytan, and Anna Goldenberg.
    \newblock Unsupervised representation learning for time series with temporal
    neighborhood coding.
    \newblock In \emph{International Conference on Learning Representations}, 2021.
    \newblock URL \url{https://arxiv.org/abs/2106.00750}.

    \bibitem[von Ziegler et~al.(2024)von Ziegler, Roessler, Sturman, Waag,
    Privitera, Duss, O'Connor, and Bohacek]{vonziegler2024flow}
    Lukas~M. von Ziegler, Fabienne~K. Roessler, Oliver Sturman, Rebecca Waag,
    Mattia Privitera, Sian~N. Duss, Eoin~C. O'Connor, and Johannes Bohacek.
    \newblock Analysis of behavioral flow resolves latent phenotypes.
    \newblock \emph{Nature Methods}, 2024.
    \newblock \doi{10.1038/s41592-024-02500-6}.

    \bibitem[Weinreb et~al.(2024)Weinreb, Pearl, Lin, Osman, Zhang, Annapragada,
    Conlin, Hoffmann, Makowska, Gillis, Jay, Ye, Mathis, Mathis, Pereira,
    Linderman, and Datta]{Weinreb2024}
    Caleb Weinreb, Jonah~E. Pearl, Sherry Lin, Mohammed Abdal~Monium Osman, Libby
    Zhang, Sidharth Annapragada, Eli Conlin, Red Hoffmann, Sofia Makowska,
    Winthrop~F. Gillis, Maya Jay, Shaokai Ye, Alexander Mathis, Mackenzie~W.
    Mathis, Talmo Pereira, Scott~W. Linderman, and Sandeep~Robert Datta.
    \newblock Keypoint-moseq: parsing behavior by linking point tracking to pose
    dynamics.
    \newblock \emph{Nature Methods}, 21\penalty0 (7):\penalty0 1329--1339, Jul
    2024.
    \newblock ISSN 1548-7105.
    \newblock \doi{10.1038/s41592-024-02318-2}.
    \newblock URL \url{https://doi.org/10.1038/s41592-024-02318-2}.

    \bibitem[Weinreb et~al.(2026)Weinreb, Kannan, Newman-Boulle, Sainburg, Gillis,
    Plotnikoff, Makowska, Pearl, Osman, Linderman, and Datta]{weinreb2026tasks}
    Caleb Weinreb, Lakshanyaa~Thamarai Kannan, Alia Newman-Boulle, Tim Sainburg,
    Winthrop~F. Gillis, Alex Plotnikoff, Sofia Makowska, Jonah~E. Pearl, Mohammed
    Abdal~Monium Osman, Scott~W. Linderman, and Sandeep~Robert Datta.
    \newblock Spontaneous behavior is a succession of self-directed tasks.
    \newblock \emph{Neuron}, 114\penalty0 (5):\penalty0 922--937, 2026.
    \newblock \doi{10.1016/j.neuron.2025.11.021}.

    \bibitem[Wiltschko et~al.(2015)Wiltschko, Johnson, Iurilli, Peterson, Katon,
    Pashkovski, Abraira, Adams, and Datta]{Wiltschko_2015}
    Alexander~B. Wiltschko, Matthew~J. Johnson, Giuliano Iurilli, Ralph~E.
    Peterson, Jesse~M. Katon, Stan~L. Pashkovski, Victoria~E. Abraira, Ryan~P.
    Adams, and Sandeep~Robert Datta.
    \newblock Mapping sub-second structure in mouse behavior.
    \newblock \emph{Neuron}, 88\penalty0 (6):\penalty0 1121--1135, Dec 2015.
    \newblock ISSN 0896-6273.
    \newblock \doi{10.1016/j.neuron.2015.11.031}.
    \newblock URL \url{http://dx.doi.org/10.1016/j.neuron.2015.11.031}.

  \end{thebibliography}
\end{document}